\documentclass[runningheads]{llncs}

\usepackage{eccv}

\usepackage{eccvabbrv}

\usepackage{graphicx}
\usepackage{booktabs}
\usepackage{listings}

\usepackage[accsupp]{axessibility}  

\usepackage{hyperref}

\usepackage{orcidlink}
\usepackage{wrapfig}

\usepackage[normalem]{ulem}
\usepackage{xcolor}

\def \FnameFigsSuppAllRes {figs/figs-viewneti-latex-all-nvs-results-resolutiongood.pdf}

\def \FnameFigsSupp {figs/figs-viewneti-latex-supp-resolutiongood.pdf}

\begin{document}

\title{Viewpoint Textual Inversion: Discovering Scene Representations and 3D View Control in 2D Diffusion Models}

\titlerunning{Viewpoint Textual Inversion}

\author{James Burgess\inst{1}\orcidlink{0000-0002-0823-2848} \and
Kuan-Chieh  Wang\inst{2}\orcidlink{0000-0002-6785-8146} \and
Serena Yeung-Levy\inst{1}\orcidlink{0000-0003-0529-0628}}

\authorrunning{J.~Burgess et al.}

\institute{Stanford University \email{\{jmhb,syyeung\}@stanford.edu} \and
Snap Inc. \email{jwang23@snapchat.com}
}

\maketitle

\begin{abstract}
Text-to-image diffusion models generate impressive and realistic images, but do they learn to represent the 3D world from only 2D supervision? We demonstrate that yes, certain 3D scene representations are encoded in the text embedding space of models like Stable Diffusion. Our approach, Viewpoint Neural Textual Inversion (ViewNeTI), is to discover \textit{3D view tokens}; these tokens control the 3D viewpoint -- the rendering pose in a scene -- of generated images. Specifically, we train a small neural mapper to take continuous camera viewpoint parameters and predict a view token (a word embedding). This token conditions diffusion generation via cross-attention to produce images with the desired camera viewpoint. Using ViewNeTI as an evaluation tool, we report two findings: first, the text latent space has a continuous view-control manifold for particular 3D scenes; second, we find evidence for a generalized view-control manifold for all scenes. We conclude that since the view token controls the 3D `rendering' viewpoint, there is likely a scene representation embedded in frozen 2D diffusion models. Finally, we exploit the 3D scene representations for 3D vision tasks, namely, view-controlled text-to-image generation, and novel view synthesis from a single image, where our approach sets state-of-the-art for LPIPS. Code available at \url{https://github.com/jmhb0/view_neti}


\end{abstract}    

\begin{figure}
    \centering
    \includegraphics[width=0.95\linewidth, page=7, trim={0 790 465 0},clip]{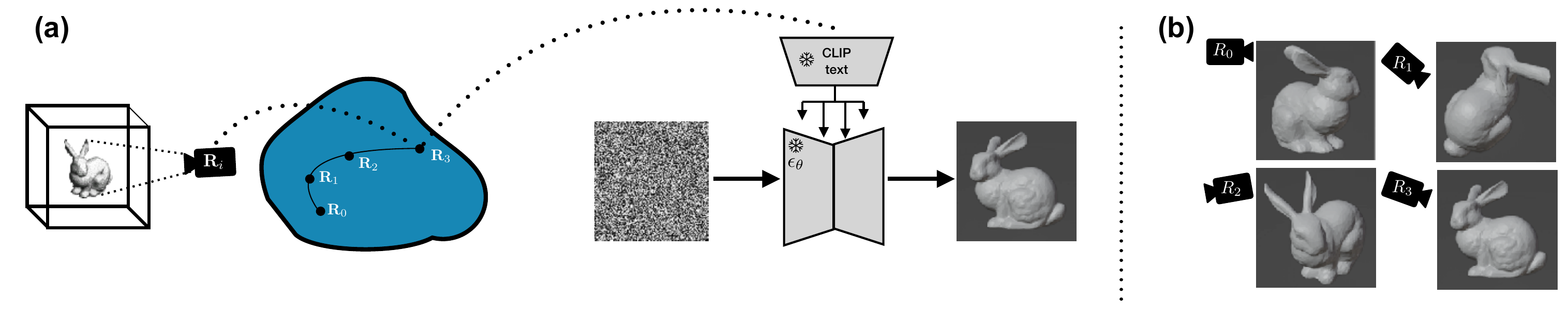}
    \captionof{figure}{We find `3D view tokens’ in the Stable Diffusion word embedding space. (a) Given a camera pose, we predict a token (word embedding), which we use to condition diffusion generation. 
    (b) Different view tokens give different views of the generated 3D scene. 
    We use 3D view tokens to study scene representations in diffusion models.
   }
    \label{fig:pull-figure}
\end{figure}

\section{Introduction}
\label{sec:intro}
\begin{figure}[h]
    \centering
    \includegraphics[width=0.7\columnwidth,page=2, trim={0 673 520 0},clip]{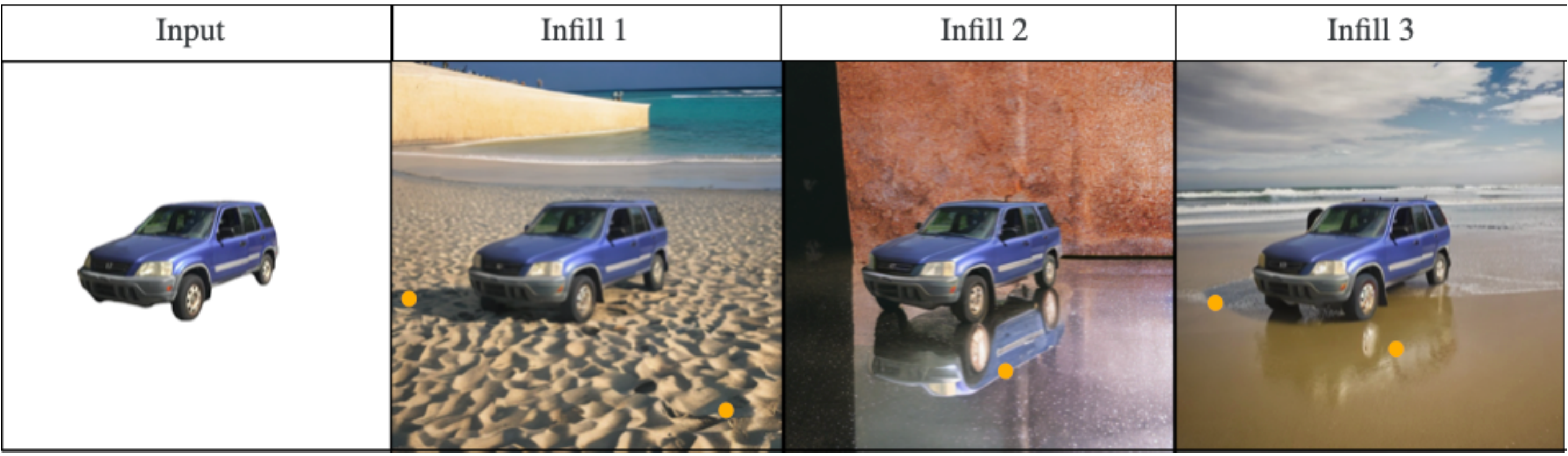}
    \caption{A masked-out car (left) with infilling (images 2 to 4) by a Stable Diffusion model \cite{stable_diffusion_rombach2022high}, with important details marked with {\color{orange} orange dots}. Infill image 1 has  shadows that are consistent with the shadows on the car. Infill image 2 has object reflections. Infill image 3 has reflections and shadows. This is evidence that 2D diffusion models are capable of 3D reasoning, which motivates our investigation into 3D view control.}
    \label{fig:car-infilling}
\end{figure}

Text-to-image diffusion models have impressive capabilities in reasoning about objects, the composition of multiple objects, and 2D spatial layout  \cite{stable_diffusion_rombach2022high, diffusion_og_sohl2015deep, og_ddpm_ho2020denoising, og_scorematching_song2019generative, saharia2022photorealistic}. Despite being trained on 2D images, they seem able to do 3D reasoning: in a simple experiment, we ask Stable Diffusion  \cite{stable_diffusion_rombach2022high} to infill the background around a car and find that it generates 3D-consistent shadows and reflections (\cref{fig:car-infilling}). This suggests that diffusion models may contain an internal 3D model of scenes that they implicitly `render'. In this work, we show evidence for such a 3D scene representation.

\begin{wrapfigure}{r}{0.5\textwidth} 
    \centering
    \includegraphics[width=0.48\textwidth, page=1, trim={597 620 808 130},clip]{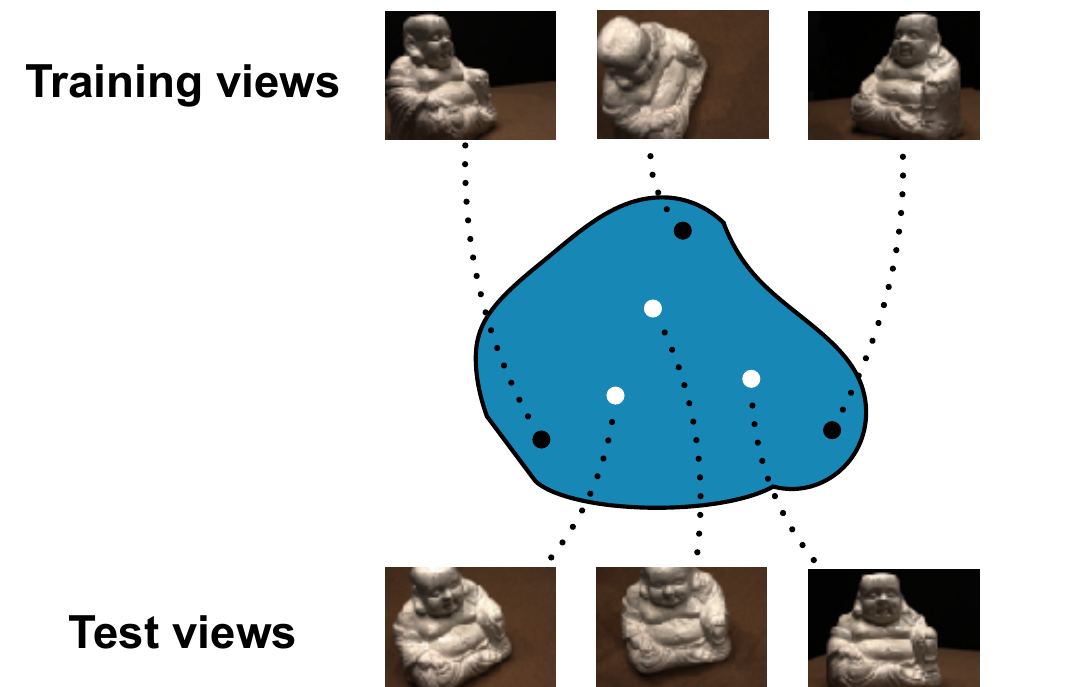}
    \captionof{figure}{We find a \textit{continuous view-control manifold} in word embedding space for one scene, by learning a token from a few training views that generalizes to test views.
   }
   \vspace{-2em}
    \label{fig:single-scene-concept}
\end{wrapfigure}

Our key insight is that the 3D viewpoint of generated images can be controlled in the word embedding space through a `3D view token'. We propose a method to discover this control mechanism called Viewpoint Neural Textual Inversion (ViewNeTI). Specifically, we learn a small neural mapping network that takes camera parameters and predicts a word embedding (the view token) to be added to a text prompt; the text prompt then conditions the diffusion via text encoding and cross-attention to produce images with the correct camera view (\cref{fig:pull-figure}). The mapper is trained using textual inversion (TI) \cite{ textual_inversion_gal2022, neti_alaluf2023neural} on very small datasets of posed images. We use ViewNeTI to evaluate the 3D control in frozen diffusion models in two settings to show two key results. 
The first result finds a continuous view-control manifold for a particular scene; the second result shows evidence for a view control manifold that is general and works for all scenes.

For our first key result, we show the existence of a \textit{continuous view-control manifold} in the text embedding space. As in ~\cref{fig:single-scene-concept}, we train ViewNeTI on a single scene with as few as three posed images. Then, we can pass new poses to ViewNeTI to move along the word-space manifold, and we can use those words as a prompt to generate novel views; the view-control token \textit{generalizes to new viewpoints}. Importantly, ViewNeTI is trained with only a handful of posed images (3-9), with a very small mapper network (140k params), and without changing the underlying diffusion model parameters. This gives us confidence that we \textit{find} a pre-existing 3D control mechanism in the text embedding space, rather than learning a new control mechanism. Since this single token can change 3D `rendering' viewpoint while keeping consistent semantics, we argue that there is likely a scene representation embedded in frozen diffusion models. 

The view-control manifold found in the single-scene setting has two limitations. The first limitation is only relevant to applications:  ViewNeTI can interpolate between training viewpoints, but not extrapolate. The second --- more important  --- limitation is relevant to understanding scene representations: the view token does not generalize to new scenes, and is entangled with the semantics of the training scene. This motivates a second experiment towards finding a more general view-control token.

\begin{wrapfigure}{r}{0.5\textwidth} 
    \centering
    \includegraphics[width=0.48\textwidth, page=1, trim={1130 552 206 82},clip]{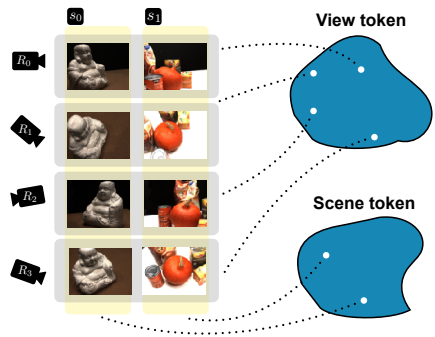}
      \captionof{figure}{Evidence for a \textit{semantically disentangled view-control manifold}. Each scene (columns) maps to a scene token, while each view (rows) maps to a view token that is shared across scenes. 
      }
    \label{fig:multi-scene-concept}
\end{wrapfigure}

For our second key result, we provide evidence for a \textit{semantically-disentangled view control manifold} in the word embedding space. 
As in ~\cref{fig:multi-scene-concept}, we jointly optimize a single ViewNeTI mapper over multiple ($<\!\!100$) scenes. 
For each scene (the columns of ~\cref{fig:pull-figure}c), we learn a separate `scene token': it should capture scene-specific semantics disentangled from viewpoint.
The view token --- the jointly-optimized ViewNeTI mapper --- is shared across scenes (the rows of ~\cref{fig:multi-scene-concept}). The view token  \textit{somewhat generalizes to new scenes}, which we demonstrate through two applications. First, we compose the learned view token with text prompts for text-to-image generation. The view token can control viewpoint on new scenes, however there is some entanglement of image style with the training dataset. Second, we use the view token for the very challenging task of novel view synthesis (NVS) from a single image.
Multi-scene training addresses the two limitations of the single-scene training case: the view token now extrapolates outside the single-image training view; and the view control mechanism generalizes to new scenes. 
Overall, this is evidence that the Stable Diffusion word embedding space has a general view-control manifold that can act on 3D scene representations for all scenes;
and as argued in the single-scene case, this implies the existence of an internal 3D scene representation. 

The ViewNeTI framework contributes to multiple lines of work in 2D generative models. Many papers observe that diffusion model representations embed certain 2D structure that is not directly supervised, such as pixel-localized semantics \cite{tang2024emergent, luo2024diffusion, zhang2024tale, hedlin2024unsupervised} and word-to-object localization \cite{hertz2022prompttoprompt, epstein2024diffusion}. Extending this, we show that diffusion models embed 3D knowledge that was not supervised, which has been separately studied in other works from different perspectives \cite{zhan2023does, chen2023beyond, el2024probing}. As the community moves towards text-to-video models \cite{ho2022imagen}, there will be greater interest in whether they `understand' the physical 3D world \cite{sora_blog_2024}, and frameworks like ours can be adapted to shed light on such questions. Although we do not focus on  applications, ViewNeTI is promising for multiple 3D vision tasks. We are the first to show that textual inversion (TI) can be used to learn a 3D control token \cite{textual_inversion_gal2022, neti_alaluf2023neural}. For controlled text-to-image generation, ViewNeTI can control 3D viewpoint without modifying the base model, which allows portability and composability with other adapters \cite{hu2021lora, mou2023t2i, zhang2023adding_controlnet}. For single-image novel view synthesis, ViewNeTI produces views with photorealistic details for real-world objects, and by leveraging 2D Stable Diffusion as a prior, it can work with very small 3D pre-training datasets (<100 scenes).

Our key contributions are:
\begin{itemize}
    \item We propose ViewNeTI, a method for investigating 3D scene representations in frozen text-to-image diffusion models. ViewNeTI learns a `3D view token' in the word embedding space that controls the rendering view of a 3D scene.
    \item Our first experiment shows that the Stable Diffusion text space has a continuous view control manifold, at least when learned for a particular 3D scene. This implies the existence of a 3D scene representation. 
    \item Our second experiment shows evidence that the Stable Diffusion text space has a general view control manifold that is disentangled from any particular scene's content. 
    \item We present applications to view control tasks. We can control the 3D viewpoint in scenes for text-to-image generation. We solve single-view novel view synthesis, and our method has excellent image photorealism, achieving state-of-the art LPIPS on DTU \cite{jensen2014large}. 
\end{itemize}

\section{Related Work}
\label{sec:related-work}
\subsection{3D Representations in 2D Diffusion Models} 
A number of prior works directly study the implicit 3D representations in Stable Diffusion \cite{zhan2023does, chen2023beyond, el2024probing}. Like us, they ask whether diffusion models ``simply memorize superficial correlations between pixel values and words?  Or are they learning something deeper (?)'' \cite{chen2023beyond}. They differ from ours in two ways. First, their methodologies use linear probing on representations, while ours is a `proof by construction': we find a mechanism for controlling 3D viewpoint of generated images. Second, we find 3D structure in the word embedding space, while theirs studies the UNet feature maps. Our finding that 3D view is controllable with cross-attention coheres with the well-known Prompt2Prompt method that shows that 2D scene layout is controllable with cross-attention  \cite{hertz2022prompttoprompt}.

Many works use pretrained 2D diffusion models as a prior in 3D vision tasks. For example in content creation, the SDS loss uses the diffusion model as a regularizer for 2D projections of 3D models \cite{poole2022dreamfusion, wang2023score_scorejacobian}. Others fine tune Stable Diffusion to do novel view prediction \cite{liu2023zero1to3, sargent2023zeronvs}.

Text-to-image generation models are increasingly producing qualitatively realistic images with 3D interactions like lighting and reflection \cite{stable_diffusion_rombach2022high, diffusion_og_sohl2015deep, og_ddpm_ho2020denoising, og_scorematching_song2019generative, saharia2022photorealistic, podell2023sdxl}. \cite{sarkar2023shadows} studies the geometric accuracy of these effects. Similarly, video generation diffusion models  --- also trained on 2D data ---  are improving, and seem to produce videos with some 3D world consistency \cite{ho2022imagen, sora_blog_2024}.

\subsection{Textual Inversion of Diffusion Models}
Personalization aims to inject novel concepts ---  like objects and styles --- into the diffusion model vocabulary using a few image examples of that concept. Textual inversion (TI) is a popular approach that optimizes a word embedding for each new concept \cite{textual_inversion_gal2022}. 
Extensions improve the quality and editability of the learned concepts by training different text embeddings depending on the noising timestep and UNet layer \cite{voynov2023p+, neti_alaluf2023neural, zhang2023prospect}. These advances are combined in the recent NeTI model \cite{neti_alaluf2023neural}, which is the current state of the art in Textual Inversion, and we use some of their architectural ideas.

Our work is the first to use textual inversion to learn concept for 3D viewpoint and to leverage TI for novel view synthesis. A concurrent work on `Continuous 3D words' does 3D control of synthetic 3D assets \cite{cheng2024learning}. Unlike ours, they do not study the `single scene optimization' setting --- the experiment that supports conclusions about a continuous control manifold. Also, they finetune model weights, and they do not explore NVS applications.
In ~\cref{sec:multi-scene-optimization}, we optimize multiple TI tokens in a single prompt, which has only been used by Break-a-Scene \cite{avrahami2023break}. 
Another work adds camera control to standard model personalization \cite{cheng2024learning}.

\subsection{3D Control Applications} 
The first downstream application is controlling the 3D viewpoint of generated images. Many methods aim to control the properties of generated images  \cite{meng2021sdedit, zhang2023adding_controlnet, mou2023t2i, hertz2022prompttoprompt}, but they are often imprecise or require large training datasets, while Textual Inversion (TI) is a data-efficient alternative. One could further improve 3D view control with personalization methods that fine-tune the model weights \cite{ruiz2023dreambooth, kumari2023multi_customdiffusion}.  This could improve view-controlled generation, but it is out of scope: our focus is on understanding 3D scene representations in existing models without any finetuning. Still, since ViewNeTI uses TI, it has the advantage of low storage cost and is composable with finetuned model checkpoints \cite{hu2021lora}.

The second downstream application is novel view synthesis (NVS) with few training views --- even from one training view. Most approaches use an explicit 3D representation, like a NeRF \cite{wang2021nerf}. To address the challenge of sparse input views, it's common to add regularizers to novel views \cite{niemeyer2022regnerf, jain2021_dietnerf, xu2022sinnerf, wang2023sparsenerf, deng2022depth_dsnerf, roessle2022dense, seo2023let, yu2022monosdf, wynn2023diffusionerf}, modify the training process \cite{yang2023freenerf, seo2023mixnerf, truong2023sparf}, or condition the NeRF on image features that are derived from pretraining on multi-view datasets \cite{yu2021pixelnerf, lin2023vision_visionnerf, chen2021mvsnerf, chibane2021stereo}. Only a few models attempt single-image NVS \cite{deng2023nerdi, liu2023zero1to3, melas2023realfusion, xu2022neurallift, sargent2023zeronvs} by leveraging diffusion models as a data prior over 2D renders (via the  Score Distillation Loss \cite{wang2023score_scorejacobian, poole2022dreamfusion}). A smaller line of work does NVS with implicit models (`geometry free', without NeRFs) \cite{sun2018multi, tatarchenko2016multi}, and some use diffusion models \cite{liu2023syncdreamer, tewari2023diffusion, szymanowicz2023viewset, zhou2023sparsefusion, karnewar2023holodiffusion, anciukevivcius2023renderdiffusion, yoo2023dreamsparse, chan2023generative}. These approaches require pretraining on large multi-view datasets \cite{deitke2023objaverse, chang2015shapenet, reizenstein2021common}, and they usually test on images that are in the same class and covariate distribution as the training set. Compared to existing NVS methods, our approach has no explicit 3D representation, no geometric priors, no regularizers, and does not require large 3D training datasets.

\section{Background}
\subsection{Text-to-image Latent Diffusion Models}
We apply viewpoint textual inversion to text-to-image Stable-Diffusion (SD). SD's are Latent Diffusion Models (LDMs) for image generation \cite{stable_diffusion_rombach2022high} and are  trained on web-scale datasets of text-image pairs $(\mathbf{x},y)\!\sim\!\mathcal{D}$ \cite{schuhmann2022laion}. There are two components. First, a variational autoencoder (VAE) \cite{vae_kingma2013auto, vae2_rezende2014stochastic} with encoder $\mathcal{E}$ and decoder $\mathcal{D}$ compresses RGB images, $\mathbf{x}\in\mathbb{R}^{H\times W\times 3}$ to a lower-dimensional latent $\mathbf{z}_0
\!=\!\mathcal{E}(\mathbf{x})\in\mathbb{R}^{h\times w\times c}$. Second, a conditional diffusion model \cite{diffusion_og_sohl2015deep, og_ddpm_ho2020denoising, og_scorematching_song2019generative, classifier_free_guidance_ho2022classifier} is trained to generate this distribution of latents with respect to the text prompt, $y$, as $p(\mathbf{z}|y)$. 

LDMs model a diffusion process: noise is incrementally added to the latent over $T$ steps; the intermediate latents are $\mathbf{z}_t$ with $t\in[0,T]$. We learn a neural network $\epsilon_\theta$ that reverses each step by predicting the applied noise. To train this network, we simulate $\mathbf{z}_t$ by sampling isotropic Gaussian noise, $\mathbf{\epsilon}$, scaling it according to a parameter $t\!\sim\mathcal{U}(1,T)$, and adding it to $\mathbf{z}$. The training objective is for $\epsilon_\theta$ to predict $\mathbf{\epsilon}$ conditioned on noising step, $t$ and the text, $y$:

\begin{equation}
  L_{LDM} := \mathbb{E}_{ (\mathbf{x},y)\sim \mathcal{D}, \mathbf{\epsilon}\sim\mathcal{N}(0,\mathbf{I}),t\sim\mathcal{U}(1,T)} \Bigl[ \| \mathbf{\epsilon} - \epsilon_\theta(\mathbf{z}_t, y, t) \| \Bigl], 
  \label{eq:ldm_objective}
\end{equation}

One can sample from $p(\mathbf{z}|y)$ with $\mathbf{z}_T\!\sim\!\mathcal{N}(0,\mathbf{I})$, and using $\epsilon_\theta$ to run the reverse process over $T$ steps \cite{diffusion_og_sohl2015deep}, . This gives a latent sample $\tilde{z}$ which is decoded to an image, $\tilde{\mathbf{x}}\!=\!\mathcal{D}(\tilde{\mathbf{z}})$. The $\epsilon_\theta$ architecture is a conditional UNet \cite{og_unet_ronneberger2015u}. $\mathbf{z}_t$ is passed through the main UNet stream. The text is passed through a pretrained CLIP \cite{clip_radford2021learning} text encoder, giving a $d$-dim conditioning vector for each token, $\mathbf{c}(y)\in\mathbb{R}^{d \times 77}$, which is mixed with each UNet layer via cross-attention \cite{stable_diffusion_rombach2022high}.

\subsection{Textual Inversion}
\label{sec:background-ti}
In textual inversion (TI) \cite{textual_inversion_gal2022}, we learn a new word embedding (word token), $\mathbf{v}_{S_o}$, for pseudo-word $S_o$, which represents a new concept from a small dataset. The dataset,  $(\mathbf{x},y)\!\sim\!\mathcal{D}_{TI}$ contains images, $\mathbf{x}$, of that concept;  and paired captions, $y$, that are ``A photo of $S_o$''. To learn the word embedding $\mathbf{v}_{S_o}$,  we use the LDM loss as in \cref{eq:ldm_objective}, but replacing $\mathcal{D}$ with $\mathcal{D}_{TI}$, and optimizing \textit{only} with respect to $\mathbf{v}_{S_o}$. Importantly, the diffusion model weights are frozen.

A recent work proposes the NeTI model \cite{neti_alaluf2023neural}, which includes many recent advances in textual inversion \cite{zhang2023inversion, valevski2023face0, gal2023encoder, wei2023elite}. Instead of learning a single embedding for $S_o$, it predicts an embedding for each UNet cross-attention layer, $\ell$ \cite{voynov2023p+} and for each noise step, $t$ \cite{gal2023encoder_ti_t_conditioning}; this representation  space is denoted $\mathcal{P}^*$ \cite{neti_alaluf2023neural}. NeTI is implemented as a small neural network mapper, $\mathcal{M}$ conditioned on $t$ and $l$. The optimization of \cref{eq:ldm_objective} is with respect to the weights of $\mathcal{M}$ \cite{ neti_alaluf2023neural}. Our work, ViewNeTI, extends the NeTI architecture.

\section{Method}
\label{sec:method}
\begin{figure*}[t]
  \centering
  \begin{subfigure}{\linewidth}
\includegraphics[width=0.9\columnwidth,page=3, trim={0 435 596 0},clip]{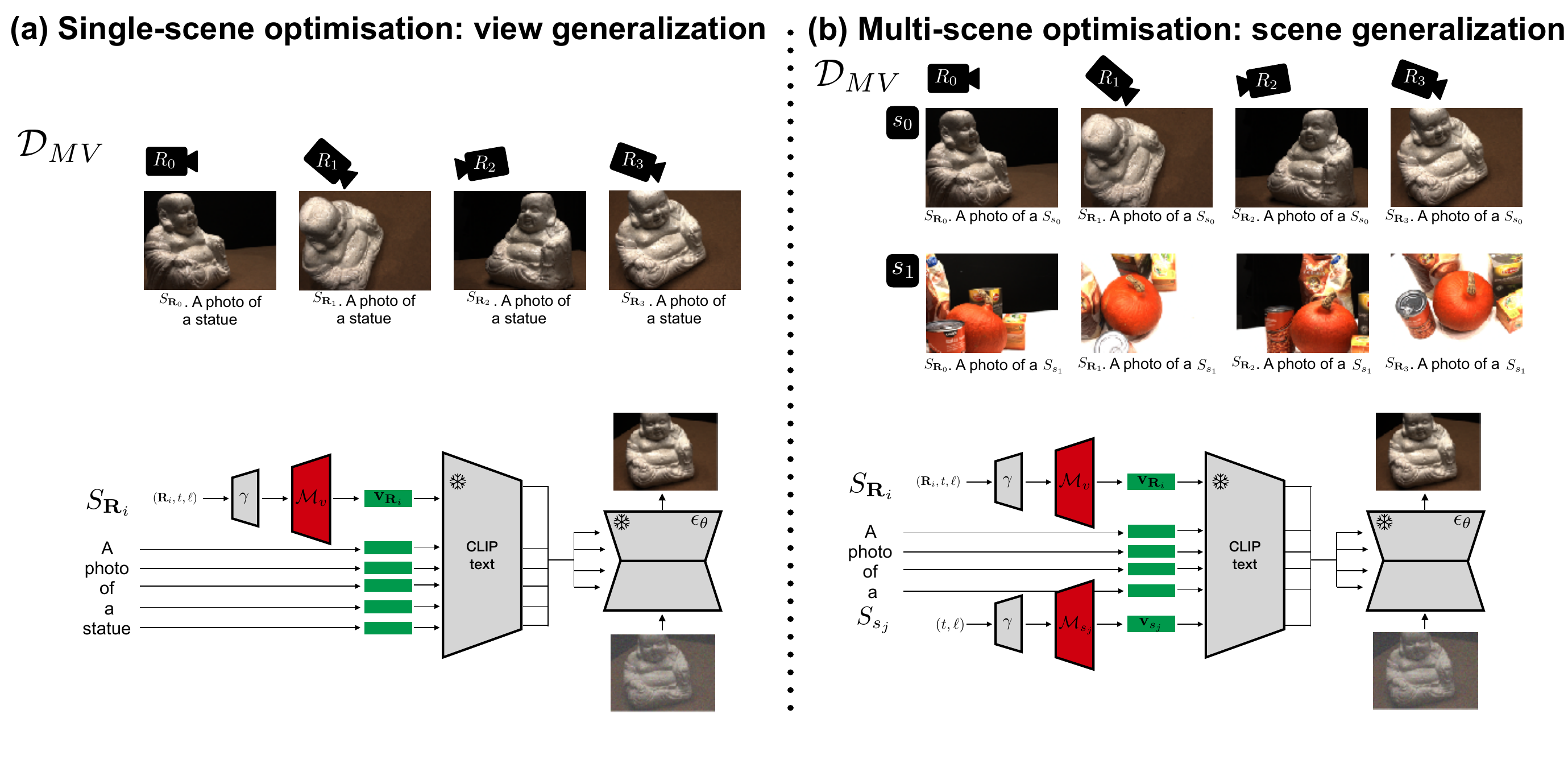}
  \end{subfigure}
  \caption{Training procedure for the `3D view token' in Viewpoint Neural Textual Inversion (ViewNeTI), our method for evaluating 3D representations in the word embedding space of frozen diffusion models. (a) To optimize a single scene (\cref{sec:single-scene-optimization}), we have (top) a small multi-view dataset, $\mathcal{D}_{MV}$ with images, $\mathbf{x}_i$, and camera poses, $\mathbf{R}_i$. We create a caption for each image, with a token $S_{\mathbf{R}_i}$ for each view, $\mathbf{R}_i$. Bottom: the embedding for $S_{\mathbf{R}_i}$ is $\mathbf{v_{\mathbf{R}_i}}$ and is predicted with a neural network $\mathcal{M}_v$, conditioned on  camera parameters, $\mathbf{R}_i$, as well as the diffusion timestep $t$, and UNet layer $\ell$. All parameters are encoded by a Fourier feature mapper, $\gamma$ \cite{tancik2020fourier2}. The other tokens take their regular word embeddings. The prompt is passed to the CLIP text encoder \cite{clip_radford2021learning}, then the text embedding is passed to the UNet via cross-attention \cite{stable_diffusion_rombach2022high}.  We do diffusion model training on this dataset while optimizing only  $\mathcal{M}_v$ (this is textual inversion training \cite{textual_inversion_gal2022, neti_alaluf2023neural}). (b) To optimize multiple scenes (\cref{sec:multi-scene-optimization}), we have a multi-view dataset with multiple scenes but shared camera poses $\mathbf{R}_i$. The optimization is the same, except each scene, $s_j$, has its own scene token $S_{s_j}$ in the caption. The view tokens, $S_{\mathbf{R}_i}$ are shared over the scenes. The embedding for $S_{s_j}$ is $\mathbf{v}_{s_j}$ and is predicted by a scene-mapper, $\mathcal{M}_{s_j}$, conditioned on timestep, $t$ and UNet layer, $\ell$. The $\mathcal{M}_v$ and $\mathcal{M}_{s_j}$ are jointly optimized.
  }
  \label{fig:sysfig}
\end{figure*}

Viewpoint Neural Textual Inversion (ViewNeTI) learns a `3D view token' that controls the viewpoint of objects in images generated by diffusion models; we use it to evaluate the latent 3D representations in Stable Diffusion. We have two training modes, depicted in \cref{fig:sysfig}. 
 
In \cref{sec:single-scene-optimization} we introduce single-scene optimization (\cref{fig:sysfig}a). We first explain the architecture and training details of the neural mapper that predicts the view token; then we cover the inference procedure for evaluating view generalization. Next in \cref{sec:multi-scene-optimization} on multi-scene optimization, we describe a pretraining strategy for learning a view token that generalizes to many scenes (\cref{fig:sysfig}b). To evaluate the cross-scene generalization, we introduce two applications: 3D controlled text-to-image generation, and single-image novel vew synthesis (NVS). 

\subsection{Single-scene Optimization and Viewpoint Generalization}
\label{sec:single-scene-optimization}
As in \cref{fig:sysfig}a, the $\mathcal{M}_v$ network predicts the 3D view token, and we optimize it from a small dataset of posed images on a single scene. This section explains the camera representation, architecture, training strategy, and inference, which allow evaluation of a continuous viewpoint control manifold in Stable Diffusion. 

\textbf{3D View / Pose Representation} The view-mapper, $\mathcal{M}_v$, is conditioned on camera parameters, denoted $\mathbf{R}_i$, for pose $i$, which can be any vector representation of the camera extrinsics and intrinsics. In our experiments, we use the camera-to-world projection matrix, and we normalize each matrix entry to the range $[-1,1]$. 
Our method is agnostic to the camera parameterization, and we verify that our method also works with spherical coordinates in ~\cref{sec:appendix-spherical-coords}.

The camera parameters, $\mathbf{R}_i$, are passed through a Fourier-feature encoding  \cite{tancik2020fourier2, rahimi2007random_fourier1} with bandwidth $\sigma=0.5$, which is necessary for the neural mappers to learn a predictor that is sufficiently sensitive to small changes in camera parameters; that is, it can represent high frequency changes in word embedding space. The $\sigma$ parameter is fixed across our experiments: big enough to model a diverse viewpoint range, but small enough to interpolate views (see ~\cref{sec:ablations} ablations). 

\textbf{ViewNeTI Mapper Architecture}
The $\mathcal{M}_v$ mapper is also conditioned on the denoising timestep and  diffusion UNet layer, $(t,\ell)$. This improves textual inversion reconstruction quality and optimization convergence because different timesteps and UNet layers control different image features; for example, small $t$ denoising steps control finer texture details rather than layout and shape\cite{neti_alaluf2023neural, voynov2023p+}. The $(t,\ell) $ parameters are also passed through Fourier encoding with bandwidths $\sigma=(0.03,2)$ that are fixed for all experiments. Formally, let the Fourier feature encoding function \cite{rahimi2007random_fourier1, tancik2020fourier2} be $\gamma(\cdot)$. We concatenate the conditioning parameters and pass them through the fourier encoding, $\mathbf{c}_\gamma=\gamma([t,\ell,\mathbf{R}])$. We fix the encoding dimension to 64. The  3D view token is then predicted by the view-mapper:
\begin{equation}
\mathbf{v}_{\mathbf{R}} =  \mathcal{M}_v(\mathbf{c}_\gamma). 
\end{equation}

The output, $\mathbf{v}_{\mathbf{R}}$, has the same dimension as the text encoder, which is 768 in SD2 \cite{stable_diffusion_rombach2022high}. We parameterize $\mathcal{M}_v$ as a 2-layer MLP with 64 dimensions, LayerNorm \cite{ba2016layer}, and LeakyRelu \cite{xu2015empirical}, and it has 140,000 parameters. Finally, we scale the embedding, $\mathbf{v}_{\mathbf{R}}$, to have the same $L_2$ norm as the word embedding for the word, `object' (the choice of this word is not important in practice).

\textbf{Single-Scene Training}
We have a small ($<10$) dataset called $\mathcal{D}_{MV}$ with multi-view images, $\mathbf{x}_i$, with known camera pose parameters, $\mathbf{R}_i$. We do not have prior 3D supervision from other multi-view datasets. As in ~\cref{fig:sysfig}a, we generate  captions of the form $y(S_{\mathbf{R}_i})=$``$S_{\mathbf{R}_i}$. a photo of a <word>'', where `<word>' is manually chosen to match the scene, e.g. `statue' (the choice of this word is not important in practice). We generate the prompt embedding and replace the $y(S_{\mathbf{R}_i})$ embedding with $\mathbf{v}_{\mathbf{R}_i}$ as described earlier. The rest follows the regular Stable Diffusion forward pass: pad the prompt embedding to 77 sequence length, pass  it through the CLIP text encoder, then cross-attend that sequence with the UNet \cite{stable_diffusion_rombach2022high}.

We  optimize the weights of $\mathcal{M}_v$ with the loss in \cref{eq:ldm_objective}, except we replace $\mathcal{D}$ with $\mathcal{D}_{MV}$. Intuitively, we learn text tokens that, when conditioning diffusion model generation, reproduce training images with the correct camera pose. 

We apply simple augmentations to  images, similar to \cite{melas2023realfusion}. This helps the accuracy of TI for small training datasets, as we show in ~\cref{sec:ablations} ablations. We also do text prompt augmentations that are standard in textual inversion \cite{textual_inversion_gal2022}. See \cref{sec:appendix-data-augmentations} for details.

\textbf{Single-scene inference} To generate novel views we simply use the same prompt: ``$S_{\mathbf{R}}$. a photo of a <word>'',  except when generating the view token, $\mathbf{v}_\mathbf{R}$ for $S_{\mathbf{R}}$, we pass new camera parameters, $\mathbf{R}$ to the view-mapper, $\mathcal{M}_v$. We then run diffusion model generation with DPMSolver \cite{lu2022dpm,lu2022dpm++,ddim_song2020denoising} for 50 steps to get the final image with the viewpoint of $\mathbf{R}$. A key observation is that since we can run inference on continuous camera parameters, then $\mathcal{M}_v$ learns a continuous 3D-control manifold; we discuss this further in ~\cref{sec:results-single-scene}.

\subsection{Multi-Scene Optimization and Scene Generalization}
\label{sec:multi-scene-optimization}
As in \cref{fig:sysfig}b, we pretrain the $\mathcal{M}_v$ network --- a predictor of the `view token' --- across multiple scenes. This section explains the pretraining of $\mathcal{M}_v$, and its two applications:  view-controlled text-to-image-generation, and novel view synthesis (NVS). These applications allow evaluation of a generalized viewpoint control manifold in Stable Diffusion. 

\textbf{Pre-training} Now we have images, $\mathbf{x}_{ij}$, with known camera poses, $\mathbf{R}_{ij}$, for views $i$ and scenes $j$ (\cref{fig:sysfig}b). 
The multi-view datasets should have dense coverage of the view space, which we visualize in ~\cref{sec:appendix-scene-camera-coverage}. For each scene, $s_j$, we define a scene token, $S_{s_j}$. Now we create a text prompt for each image: $y(S_{\mathbf{R}_i}, S_{s_j})=$``$S_{\mathbf{R}_i}$. a photo of a $S_{s_j}$''. Similar to the view token, each scene token, $S_{s_j}$, has a scene-mapper $\mathcal{M}_{s_j}$, which predicts a word embedding for the scene, $\mathbf{v}_{s_j}$. The scene-mapper is identical to the view-mapper, $\mathcal{M}_v$, but without camera viewpoint conditioning; to improve object reconstruction quality, we also add output bypass \cite{neti_alaluf2023neural}, explained in \cref{sec:appendix-output-bypass}. 

The training is the same as the single-scene case, except now we sample over the multi-scene dataset, and we jointly optimize a single view-mapper $\mathcal{M}_v$, with multiple scene mappers, $\mathcal{M}_{s_j}$. The learned view-mapper, $\mathcal{M}_v$, predicts a 3D view token for controlling viewpoint across multiple scenes; it projects into a semantically-disentangled view-control manifold. To test the quality of this generalization, we propose two applications. 

\textbf{Application 1: View-Controlled Text-to-Image Generation}
\label{sec:methods-pose-generation}
Content creation via text-to-image diffusion is a popular application, and one research direction seeks to control certain properties in generated images \cite{hertz2022prompttoprompt, zhang2023adding_controlnet}. The pretrained view-mapper can control the viewpoint in new objects by adding the view token to text prompts, for example ``$S_{\mathbf{R}_i}$. A brown teddy bear'', where the embedding for $S_{\mathbf{R}_i}$ is predicted by $\mathcal{M}_v$. Then, the text encoder conditions the standard diffusion generation. 

\textbf{Application 2: Novel View Synthesis} In novel view syntehsis (NVS), we have a dataset of images of a scene with known camera parameters, $(\mathbf{x}_i, \mathbf{R}_i)\sim\mathcal{D}_{MV}$. To apply ViewNeTI to NVS, we construct a prompt to match the pretraining: ``$S_{\mathbf{R}_i}$. a photo of a $S_{s_j}$'' where $S_{\mathbf{R}_i}$ is predicted by our pretrained view-mapper, $\mathcal{M}_v$, and we create a new scene mapper, $\mathcal{M}_{s_j}$, for $S_{s_j}$. Then, we simply optimize the new scene-mapper, $\mathcal{M}_{s_j}$, on this scene's images with the same objective used in pretraining, ~\cref{eq:ldm_objective}.

Our results in ~\cref{sec:results-multi-scene} focus on NVS from a single image. Here, the method is the same as the general NVS method, just using a single input image.

\section{Results}
\label{sec:results}
Our experiments investigate implicit 3D scene representations in the Stable Diffusion representation space by learning a continuous `3D view token' for controlling 3D viewpoint in generated images.  In ~\cref{sec:results-single-scene}, we examine the first case: a view token optimized for a single scene, which shows strong evidence for a \textit{continuous view control manifold}. In ~\cref{sec:results-multi-scene}, we report results of pretraining a more general view token for many scenes. To show its generalization properties, we use the view token in two applications: view-controlled text-to-image generation, and single-image novel view synthesis. They show evidence for a \textit{semantically-disentangled view control manifold}. Together, 
these view-control mechanisms effectively control the `rendering viewpoint' in image generations, and suggest the existence of an implicit 3D scene representation in diffusion models.

\begin{figure}[t]
  \centering
    \includegraphics[width=0.8\columnwidth,page=4, trim={0 550 42 0},clip]{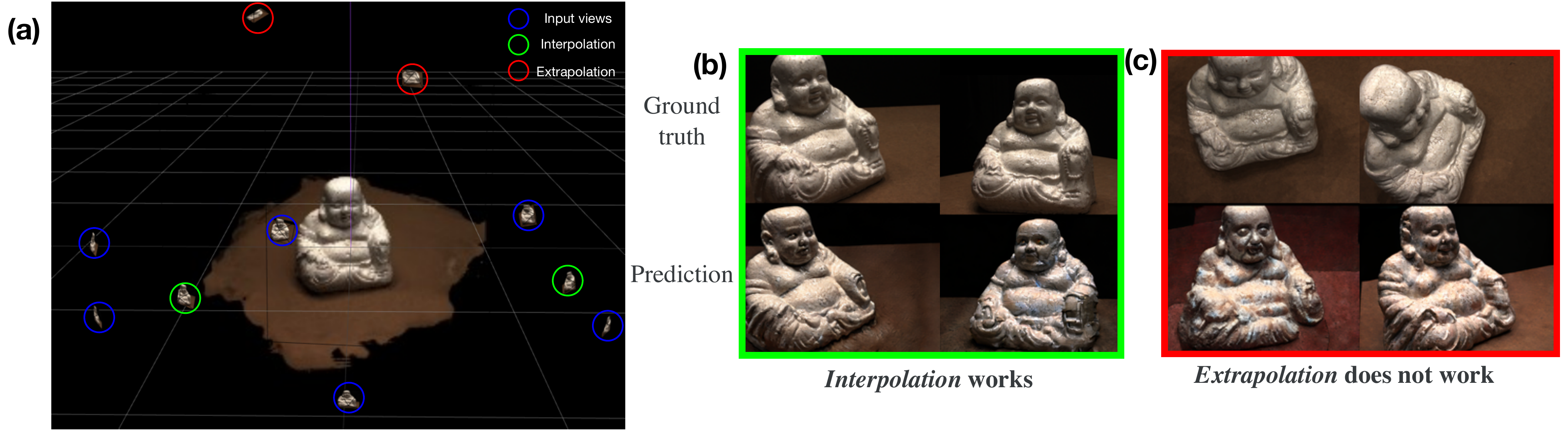}
    \centering
  \caption{Evidence for a continuous view-control manifold in Stable Diffusion by training ViewNeTI on a single scene (\cref{sec:results-single-scene}). (a) the camera positions for  DTU scene 114 \cite{jensen2014large} visualized with Nerfstudio \cite{nerfstudio} and SDFStudio \cite{Yu2022SDFStudio}. We show training views from the 6-view split (blue), inference views that are `interpolated' from the input (green), and inference views that are `extrapolated' (red). (b) the interpolated inference views are predicted correctly, showing that the word embedding space has a continuous view-control manifold. (c) inference on extrapolated views does not work; the semantics are good but the poses are wrong. This is only relevant to view-synthesis applications, and our pretraining strategy in \cref{sec:results-multi-scene} can address it.
  }
  \label{fig:results-single-scene}
  \vspace{-2em}
\end{figure}

\subsection{Single-Scene Optimization and Viewpoint Generalization}
~\label{sec:results-single-scene}
\textbf{Dataset} We evaluate on DTU \cite{jensen2014large}, a multi-view dataset of real-world objects with challenging details. We use the train-test splits for camera views used in the literature for sparse-view novel view synthesis \cite{yu2021pixelnerf, deng2023nerdi} (training set sizes 1, 3, 6, and 9). The splits are visualized in ~\cref{sec:appendix-cameras-sparse-DTU}.

\textbf{Evidence for a Continuous View Manifold} For single-scene optimization, ~\cref{fig:results-single-scene} visualizes the train and test views, as well as predictions for test views. (See ~\cref{sec:appendix_results_single_scene_optimization} for results on more scenes.). The green test views marked `interpolation' are qualitatively good novel view predictions (\cref{fig:results-single-scene}b); here `interpolation' means convex combinations of the camera parameters in spherical coordinates.  Therefore the view-mapper, $\mathcal{M}_v$, predicts a subspace in the Stable Diffusion text input space that changes the rendering viewpoint of a 3D scene with consistent semantics. The view-mapper has most likely  \textit{discovered} a pre-existing 3D control manifold in the latent space, rather than learning a new control mechanism. This is because the only 3D data the view-mapper has been trained on is six images, but it is able to \textit{generalize to new views}. Further, since this single token can change 3D `rendering' viewpoint, it is likely that a 3D scene representation exists in Stable Diffusion.

There are limitations to this finding. First --- as a minor point, and from an applications perspective --- novel view predictions do not extrapolate (\cref{fig:results-single-scene}c). Second, and more significantly, this view-control manifold is specific to a particular scene. The next results section seeks a more general view-control manifold.

\subsection{Multi-scene Optimization and Scene Generalization}
~\label{sec:results-multi-scene}
\textbf{Pretraining} We pretrain the view-mapper, $\mathcal{M}_v$, as described in ~\cref{sec:multi-scene-optimization} using the 88 train scenes from DTU chosen by \cite{yu2021pixelnerf}. The pretraining takes one day on one Titan RTX. We validate the pretraining reconstructions in ~\cref{fig:pretrain-recons}.

\begin{figure*}
  \centering
  \begin{subfigure}{\linewidth}
  \centering
    \includegraphics[width=0.9\columnwidth,page=5, trim={0 560 205 0},clip]{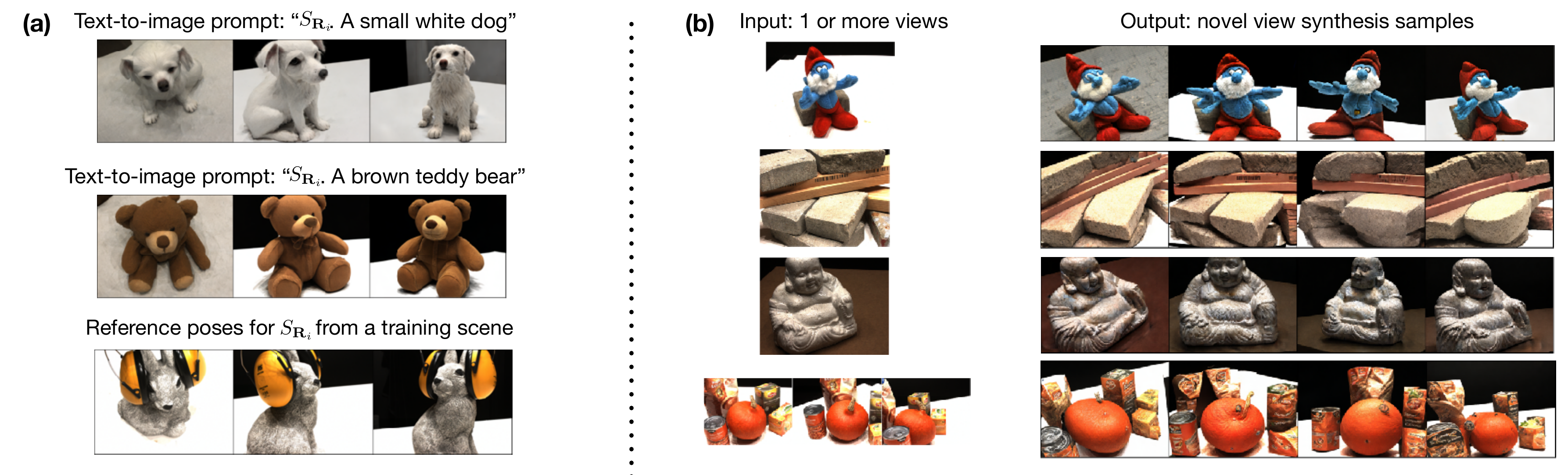}
  \end{subfigure}
  \caption{
  We train a view-token, $S_{\mathbf{R}_i}$, for controlling viewpoint in all scenes (semantic-disentanglement), and these two applications asses that generalization ability (discussion in \cref{sec:results-multi-scene}). (a) In text-to-image generation, the view token, $S_{\mathbf{R}_i}$ is added to a text prompt to generate objects not in the train set. We create prompts for three viewpoints by varying the camera parameters, $\mathbf{R}_i$ --- the columns are the same view. The bottom row has renders from a training scene: they are a reference showing how the different columns should be oriented relative to each other.
  (b) In novel view synthesis, the pretrained view token, $S_{\mathbf{R}_i}$, is used for generating novel views from few images, including from only one image. The object classes are not in the pretraining set. 
  }
  \label{fig:applications-generation-and-nvs}
  \vspace{-2em}
\end{figure*}

\textbf{View-Controlled Text-to-Image-Generation}
In ~\cref{fig:applications-generation-and-nvs}a, we show examples of using ViewNeTI to control the viewpoint of images in text-to-image generation. The object semantics are consistent across views, and using our conditioning adds negligible runtime to generation. Since the example images are outside the semantic distribution of the DTU training dataset, we claim this is evidence that the view token is at least somewhat general and semantically disentangled. Having said that, there is least some entanglement with the style of the training dataset, since generated images have similar backgrounds. 

\textbf{Single Image Novel View Synthesis}
To do single-image NVS, we take the frozen view-mapper, $\mathcal{M}_v$, and fine-tune a scene-mapper, $\mathcal{M}_{s_j}$, on a new scene (as in ~\cref{sec:multi-scene-optimization}) that has a different object class. This takes one hour on one Titan RTX. We show single-view NVS predictions for selected scenes in ~\cref{fig:applications-generation-and-nvs}b, and all DTU test scenes in Appendix ~\cref{fig:appendix-results-1nvs}. Since these new scenes are different classes, this further supports that the view token has learned a more general notion of viewpoint --- that it has discovered a semantically disentangled view-control manifold. Still, we have not tested complete generality, since the test scenes are in the same `style' as the training scenes: they have small numbers of objects with a similar background. 

Similar to the single-scene results in ~\cref{sec:results-single-scene}, we conclude that this token can change 3D `rendering' viewpoint for multiple scenes, giving further evidence that a 3D scene representation exists in Stable Diffusion. As a final point, the learned scene tokens, $S_{s_j}$ behave as 3D scene representations, since they capture the semantics of their respective scenes. 

\begin{figure*}
  \centering
  \begin{subfigure}{\linewidth}
  \centering
    \includegraphics[width=0.65\columnwidth,page=6, trim={0 350 650 0},clip]{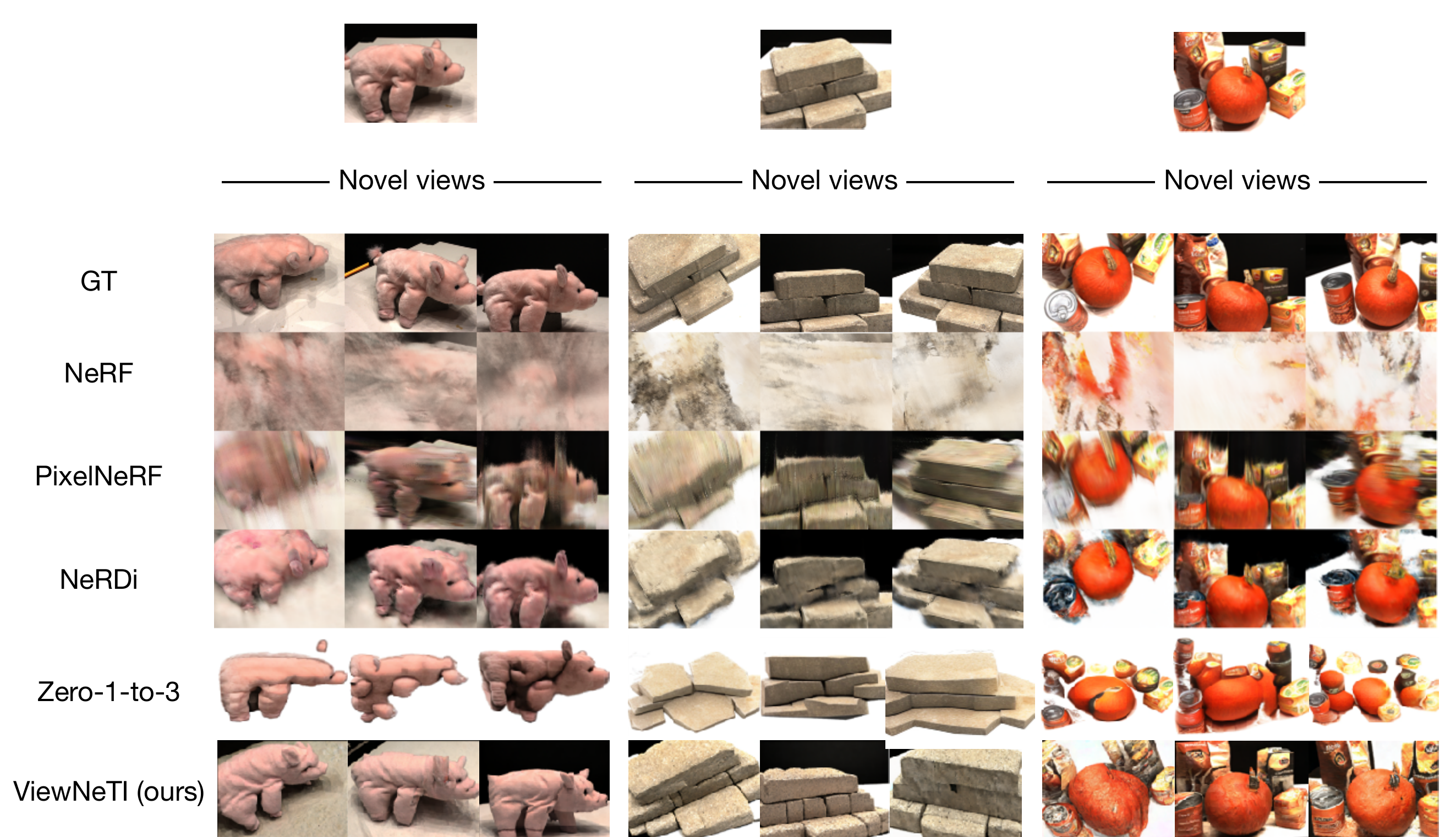}
  \end{subfigure}
  \caption{Qualitative comparison of single-view novel view synthesis on DTU \cite{jensen2014large}. Our method has good photorealism and semantics compared to baselines (see ~\cref{sec:results-multi-scene}).}
  \label{fig:baseline-comparison-single-view-nvs}
  \vspace{-2em}
\end{figure*}

\textbf{Single image Novel View Synthesis Baseline Comparisons}
Finally, we do further analysis on the single-image novel view synthesis (NVS) application, since it is a significant and challenging task. First, note that multi-scene pretraining has resolved the extrapolation issue  identified in the single-scene case in  ~\cref{fig:results-single-scene}c. In ~\cref{fig:baseline-comparison-single-view-nvs}, we compare NVS predictions against baselines using the same challenging evaluation scenes and views chosen by \cite{deng2023nerdi}.  
The first 3 methods \cite{wang2021nerf, yu2021pixelnerf, deng2023nerdi} use an explicit 3D scene representation --- a NeRF --- which ensures consistent scene geometry, but they have blurriness and artifacts that commonly appear under sparse supervision \cite{yu2021pixelnerf, niemeyer2022regnerf}, and they have errors in the semantic details. 
We also compare against Zero-1-to-3 \cite{liu2023zero1to3}, which uses a  diffusion model without explicit scene geometry, and find that it does not make reasonable predictions on DTU. This is probably because the DTU images are too different from the training distribution of centered 3D assets on white backgrounds \cite{deitke2023objaverse}. 

On the other hand, our ViewNeTI predictions, while sometimes hallucinating object details, do generate photorealistic images with plausible semantics and little blurriness. The images are highly photorealistic because they are generated by a pretrained and frozen diffusion model that we leverage as a strong prior. Moreover, ViewNeTI can generalize to the test scenes that are different classes from the pretraining scenes. Quantitatively, we compare LPIPS, SSIM, and PSNR against baselines in ~\cref{tab:results_dtu}, which is standard in NVS. Our approach is state-of-the-art for LPIPS, and near state-of-the art for SSIM, which is consistent with the strong photorealism results we see in ~\cref{fig:baseline-comparison-single-view-nvs}. Overall, ViewNeTI has strong performance in single-view NVS despite not having many components that are typically required in NVS applications: it has no large multi-view datasets, no explicit 3D representation, and no task-specific 3D regularizations. 

\begin{table}[t]
\caption{Single-image novel view synthesis metrics on DTU. The best score is in \textbf{bold}, and second best is \underline{underlined}. (Discussion in  ~\cref{sec:results-multi-scene})} 
\label{tab:results_dtu}
\centering 
\scriptsize 
\begin{tabular}{lccr}
\toprule
\textbf{Method}                                     & \textbf{LPIPS} $\downarrow$ & \textbf{SSIM} $\uparrow$ & \textbf{PSNR} $\uparrow$ \\ \midrule
NeRF \cite{wang2021nerf}                            &  0.703 & 0.286      &  8.000         \\
pixelNeRF \cite{yu2021pixelnerf}                    & 0.515  & \textbf{0.564} & \underline{16.048} \\
SinNeRF \cite{xu2022sinnerf}                        & 0.525 & \underline{0.560}  & \textbf{16.520} \\ 
NerDi \cite{deng2023nerdi}                          & \underline{0.421} & 0.465  & 14.472 \\ 
\hline
ViewNeTI (ours)                                     & \textbf{0.378}    &   0.516  &  10.947\\ 
\bottomrule
\end{tabular}
\vspace{-.5cm}
\end{table}

\subsection{Ablations}
\label{sec:ablations}
The key design decision is training the view-mapper $\mathcal{M}_v$, and we show qualitative results on ablations in \cref{sec:appendix-ablations}. This shows that the key  design choices are the frequency of the positional encoding and the text embedding norm scaling. The image augmentation strategy is also essential to avoiding degenerate solutions.

\section{Limitations and Future Work}
\label{sec:limitations-future}
Our first key result shows that Stable Diffusion has a continuous view-control manifold in the text space (\cref{sec:results-single-scene}), however our approach is only qualitative. We cannot provide metrics to, for example, compare the quality of 3D representations between different diffusion models. One approach would be to measure view synthesis reconstruction accuracy, however this may not be convincing if the optimal design parameters for ViewNeTI differ between architectures. 

Our second key result was evidence for a semantically-disentangled view-control manifold (\cref{sec:results-single-scene}), but our results show that the view-token is not fully-general. In particular, the text-to-image generations have some semantic entanglement with the style of the pre-training scenes, and the view synthesis results have some errors and hallucinations. Future work could pre-train on a larger corpus, and consider strategies like staged training, so that the object token is learned before the view token (similar to \cite{cheng2024learning}).

Our view-control tokens were learned on the DTU camera coordinates (visualized in \cref{fig:camera-positions}), which cover part of the surface of a sphere. Future work could investigate a large space of camera parameters, for example on datasets with buildings or outdoor scenes. 

Since we focus on understanding representations in frozen  models, there is significant opportunity to build on applications. Two directions just mentioned --- better disentanglement, and larger camera parameter space --- would improve the quality of both view-controlled text-to-image generation and NVS. For NVS in particular, ViewNeTI has errors in object details, such as small textures in the building scene (\cref{fig:appendix-results-1nvs}, row 2). It also makes errors in the precise object pose. Together, the PSNR is below SOTA in ~\cref{tab:results_dtu}. Reconstruction quality is an active area of research in textual inversion \cite{neti_alaluf2023neural} and advances there should be transferable to ViewNeTI. Another approach that is likely to work is full-model finetuning, for example with LORA \cite{hu2021lora}.

 
\section{Conclusions}
In this study, we discovered a 3D view-control token in Stable Diffusion; this view token controls rendering perspective in a scene, suggesting that diffusion models embed some 3D scene representation.
The question of scene understanding in diffusion models has received attention recently \cite{zhan2023does, chen2023beyond, sarkar2023shadows}, and our work provides a constructive proof, as well as the first evidence that the text space can do 3D control via cross-attention, similar to how it does 2D layout control \cite{hertz2022prompttoprompt}.

 With the release of impressive text-to-video models \cite{ho2022imagen, sora_blog_2024}, similar questions are being asked: can 2D only models learn to represent 3D geometry and physics? Approaches like ours can help to understand these questions. 

Beyond interpretability of 3D scene representations, a popular research direction is using 2D models as a prior for 3D applications \cite{poole2022dreamfusion, liu2023zero1to3}.
This is appealing because 2D data is  easier to acquire than 3D data. 
We believe that methods like ours offer a path to improving data efficiency of 3D methods by tapping into the 3D understanding implicitly learned by 2D-only models. For example, our single-image NVS application required 3D pre-training on fewer than 100 scenes, compared with competing methods like Zero-1-to-3 that use over 100,000 scenes. 

% ...

%
%
\bibliographystyle{splncs04}
\bibliography{main}

\clearpage
\appendix
\setcounter{page}{1}

\section*{Supplementary overview}
The code is submitted as available at \href{https://github.com/jmhb0/view_neti}{https://github.com/jmhb0/view\_neti}

\begin{figure*}
  \centering
    \includegraphics[width=\linewidth, page=10, trim={0 100 215 0}, clip]{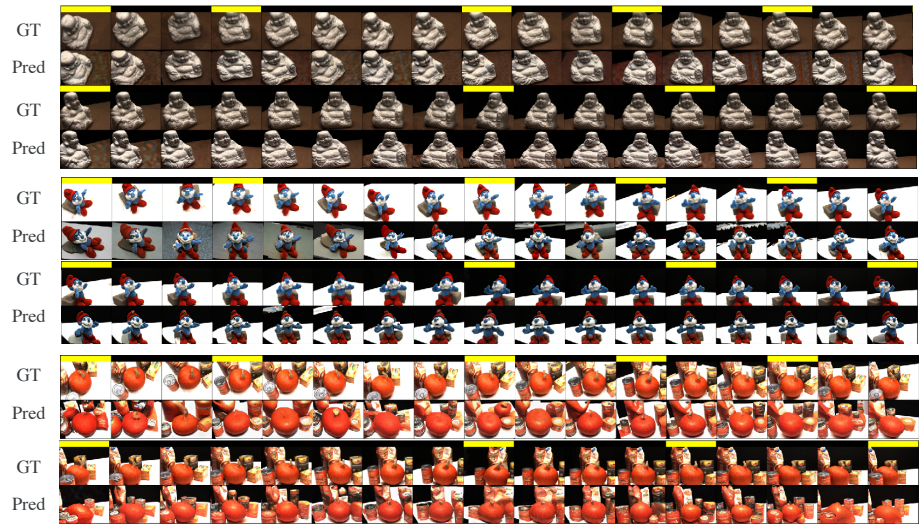}
  \captionof{figure}{
  Ground truth and predictions for single-scene optimization on 9 images for DTU scans 114, 82, and 31. The chart shows all 9 train views and all test views that are standard to evaluate on; the train views have a yellow bar. All of these scenes were trained with the same hyperparameters and sampled with the same random seed.
  }
  \label{fig:appendix-single-scene-optimization-all-results-1}
\end{figure*}

\begin{figure*}
  \centering
    \includegraphics[width=\linewidth, page=11, trim={0 100 215 0}, clip]{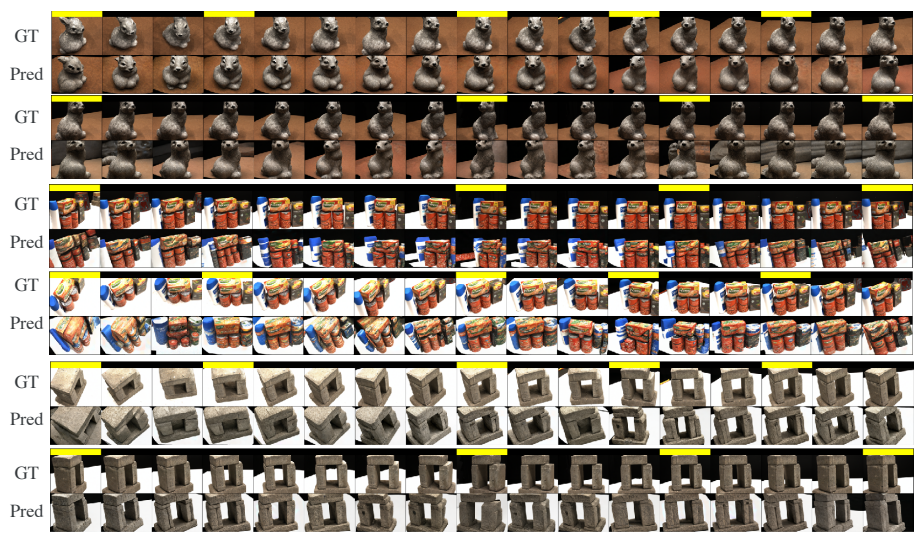}
  \captionof{figure}{
  Ground truth and predictions for single-scene optimization on 9 images for DTU scans 65, 45, and 40. The chart shows all 9 train views and all test views that are standard to evaluate on; the train views have a yellow bar. All of these scenes were trained with the same hyperparameters and sampled with the same random seed.
  }
  \label{fig:appendix-single-scene-optimization-all-results-2}
\end{figure*}

All supplementary figures are located after the supplementary text.  Especially important are  ~\cref{fig:appendix-single-scene-optimization-all-results-1} and ~\cref{fig:appendix-single-scene-optimization-all-results-2}, which are extend results from the single-scene optimization in ~\cref{sec:results-single-scene}. They show more scenes and more views. For applications, the most important figures are ~\cref{fig:appendix-results-1nvs} and ~\cref{fig:appendix-results-3nvs}, which are 3-view and single-view novel view synthesis predictions for every scene in the DTU test set. 

\begin{figure*}
  \centering
    \includegraphics[width=\linewidth, page=6, trim={0 435 550 0}, clip]{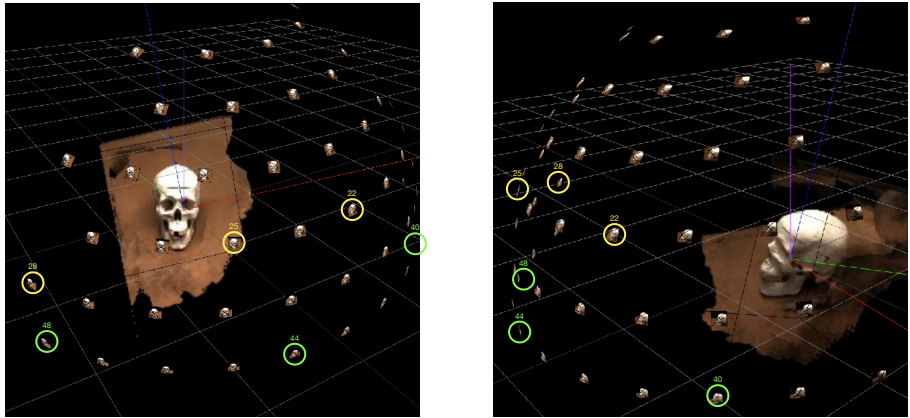}
  \captionof{figure}{Visualization of the camera view positions in a sample DTU scene \cite{jensen2014large}. The camera positions are the same for all scenes, and  the left and right images are the same 3D scene from different angles. Single-view training uses view 25. Three-view training uses the same view 25, plus views 22 and 28 (yellow). Six-view training uses those three views, plus views 40, 44, and 48 (green).  Visualization by running SDFStudio \cite{Yu2022SDFStudio}, and visualized with nerfstudio \cite{nerfstudio}.}
  \label{fig:camera-positions}
\end{figure*}

The supplementary sections and their figures are: 
\begin{itemize}
    \item A: diffusion model outfilling tests as evidence that diffusion models do 3D reasoning. ~\cref{fig:outfilling}.
    \item B: qualitative ablations of ViewNeTI design decisions for single-scene optimization. ~\cref{fig:ablations-114}, ~\cref{fig:ablations-31}.
    \item C: image and text prompt augmentation details. ~\cref{fig:augs}.
    \item D: single-scene optimization results. ~\cref{fig:appendix-single-scene-optimization-all-results-1}, ~\cref{fig:appendix-single-scene-optimization-all-results-2}.
    \item E: visualizations of the camera positions in DTU pretraining data. ~\cref{fig:camera-positions}.
    \item F: visualizations of the camera positions in DTU sparse view datasets (1, 3, and 6 views). ~\cref{fig:camera-positions}.
    \item G: validation of ViewNeTI pretraining of the view-mapper, and disentanglement of the scene-mapper. ~\cref{fig:pretrain-recons}, ~\cref{fig:disentanglement}.
    \item H: validation that ViewNeTI works with spherical coordinate camera parameterization. ~\cref{fig:spherical-coords}.
    \item I: comparison to Zero-1-to-3 \cite{liu2023zero1to3}, a baseline for single-view NVS. 
    \item J: implementation details for ViewNeTI.
    \item K: implementation details for evaluation on DTU.
    \item L: output bypass modification to the scene-mapper implementation. ~\cref{fig:output-bypass}.
    \item M: single image NVS on DTU with pretraining on Objaverse, compared to Zero-1-to-3 \cref{fig:objaverse_training}
\end{itemize}

\section{Evidence for 3D capabilities in diffusion models with image outfilling}

\label{sec:appendix-infill-experiment}
\begin{figure*}
  \centering
    \includegraphics[width=\linewidth, page=1, trim={0 420 700 0},clip]{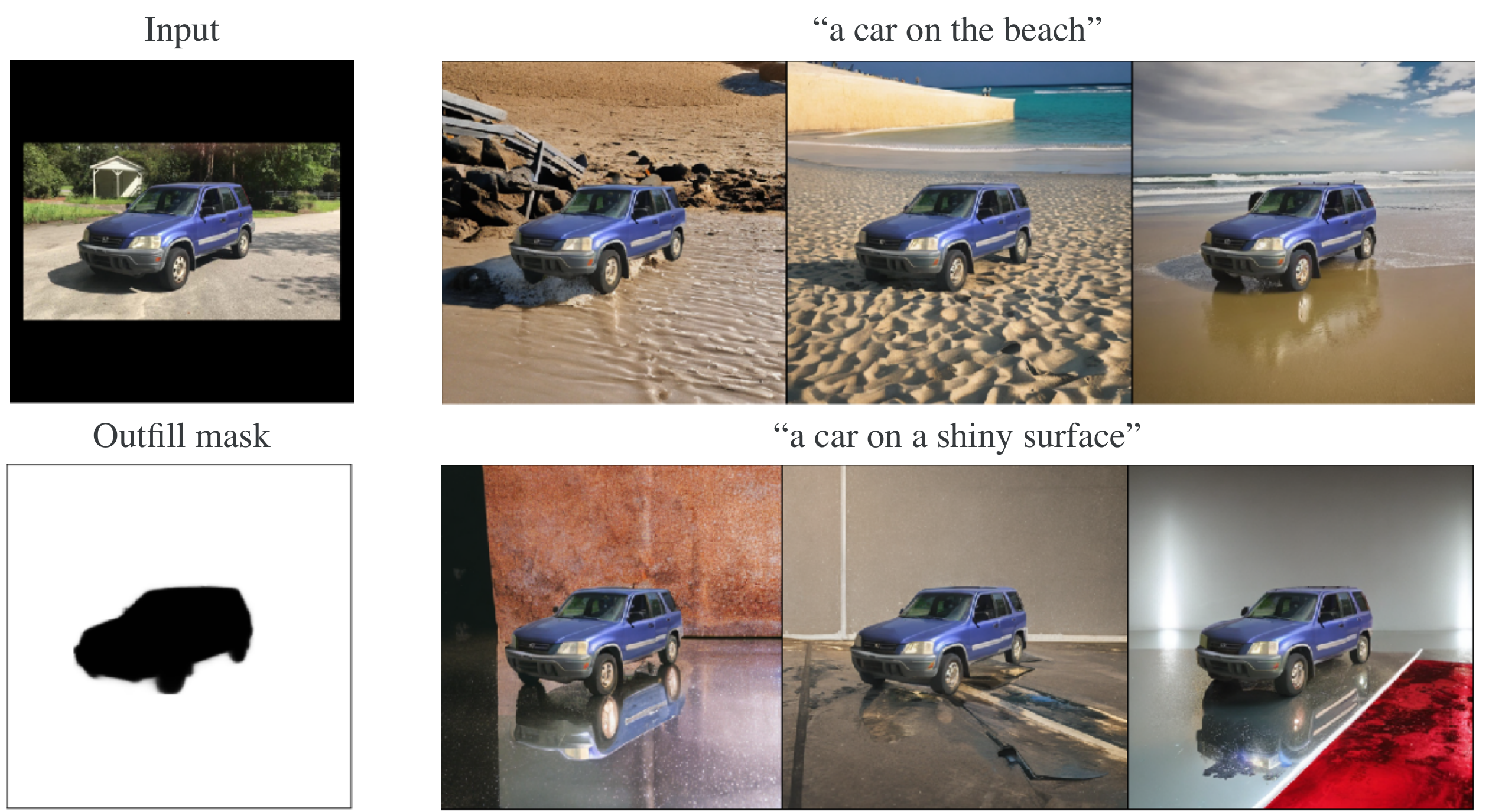}
  \captionof{figure}{Evidence that image diffusion models can reason in 3D. We take the car on the top left, and the mask in the bottom left \cite{reizenstein2021common}, and run a Stable Diffusion \cite{stable_diffusion_rombach2022high} inpainting model. The model only sees the car object itself. The shadows inferred for the `beach' prompt are the same as the shadow in the real image; the model has inferred the shadow placement from the lighting pattern on the car. For the `shiny surface' prompt, the model has reflected the car object. We also see reflections for the right-side beach scene. 
  }
  \label{fig:outfilling}
\end{figure*}

As discussed in ~\cref{sec:intro}, our work is motivated by the observation that 2D image diffusion models seem capable of reasoning about 3D phenomena. In ~\cref{fig:outfilling} we ask a diffusion model to do outfilling around a real car. In the figure caption, we discuss the evidence for 3D reasoning. The model is a Stable Diffusion,  \cite{stable_diffusion_rombach2022high} \texttt{stabilityai/stable-diffusion-2-inpainting} checkpoint, run for 50 denoising steps. The car and the mask are from Common Objects in 3D \cite{reizenstein2021common}, car id \texttt{106\_12650\_23736}.

\section{Ablations}
\label{sec:appendix-ablations}
\begin{figure*}
  \centering
    \includegraphics[width=\linewidth, page=2, trim={0 500 445 0}, clip]{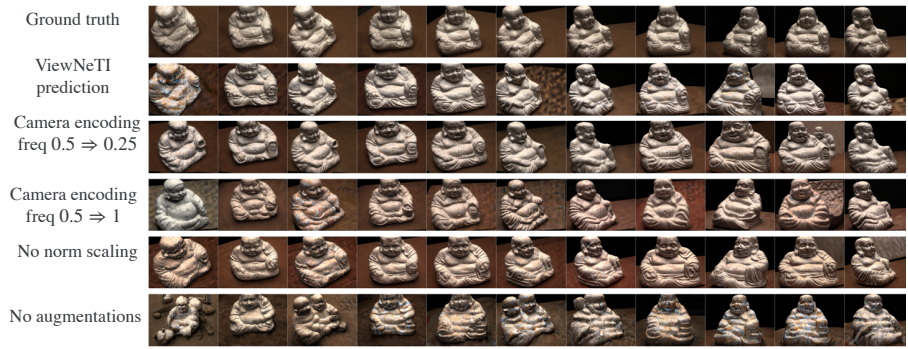}
  \captionof{figure}{Ablation of design choices for training ViewNeTI on a single scene on DTU scan114. Test view indexes are listed in ~\cref{sec:appendix-eval}.
  }
  \label{fig:ablations-114}
\end{figure*}

\begin{figure*}
  \centering
    \includegraphics[width=\linewidth, page=3, trim={0 500 445 0}, clip]{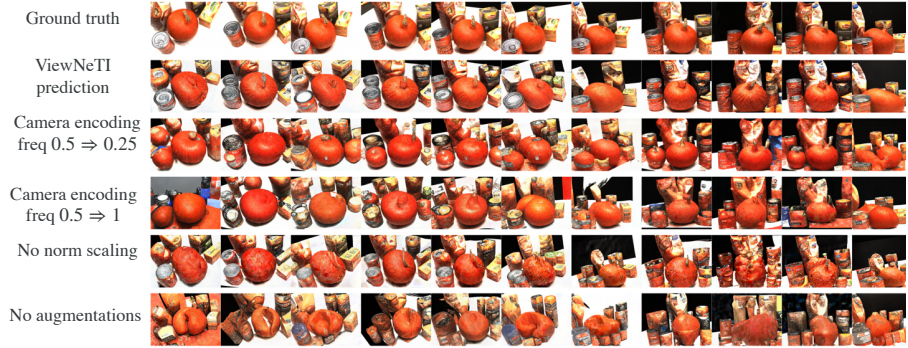}
    \captionof{figure}{
    Ablation of design choices for training ViewNeTI on a single scene on DTU scan31. Test view indexes are listed in ~\cref{sec:appendix-eval}.
  }
  \label{fig:ablations-31}
\end{figure*}

We do qualitative ablations for the design choices of ViewNeTI training on the single scene optimization case. The figures are ~\cref{fig:ablations-114} and ~\cref{fig:ablations-31}. The too-low frequency encoding for camera parameters causes difficulty in covering all the camera views. The learning of scene semantics is worse when having no augmentations, when having too-high frequency encoding, and when missing the norm scaling.

\section{Data Augmentation}
\label{sec:appendix-data-augmentations}
As in ~\cref{sec:single-scene-optimization}, we apply the image augmentations to help learn a robust scene token. The augmentations are  similar to \cite{melas2023realfusion} with some changes: no grayscaling, because that leads to some gray generations; higher probability Gaussian blur; and no horizontal flips, since it would prevent learning the view-mapper. The ablations in ~\cref{sec:appendix-ablations} demonstrate the necessity of these augmentations. For 3-view NVS, we found that such strong augmentations were not necessary, so we reduced the \texttt{size} parameter of the  \texttt{RandomResizedCrop} to \texttt{(0.950, 1.05)}. 

We also do text prompt augmentations. As described in Methods, the text encoder input is the prompt ``$S_{R_i}$. A photo of an $S_{s_j}$'', where $S_{R_i}$ and $S_{s_j}$ are controlled by the view- and scene-mappers respectively. This is the same as regular textual inversion, but with the view token \cite{textual_inversion_gal2022}. Following that work, we use different text templates, for example ``$S_{R_i}$. a rendition of the $S_{s_j}$.'' and ``$S_{R_i}$. a photo of a small $S_{s_j}$.'' \cite{textual_inversion_gal2022, clip_radford2021learning}. We use the same templates as \cite{neti_alaluf2023neural}, which are also available in our code.

\section{Single-Scene Optimization Results}
\label{sec:appendix_results_single_scene_optimization}

In ~\cref{sec:results-single-scene}, we showed results for single-scene optimization. We trained a single view-mapper on 6 views and generated test views. In ~\cref{fig:appendix-single-scene-optimization-all-results-1} and ~\cref{fig:appendix-single-scene-optimization-all-results-2} we show more scenes and more views. Each subfigure has ground truth DTU images on the top row and generated views on the bottom row. The training views have a yellow bar on the top, and the remaining views are predictions. We use all the views that are standardly used in DTU evaluation. Note that unlike in the main text, we train with 9 views, so all the test views are considered `interpolations'. The results for training with `6 views' are similar, except the extrapolated views are bad predictions. 

The idea is that for all scenes, the test views generate novel views that were not visible at training, supporting the claim that there is a continuous 3D view-control manifold in the word space.

\section{Visualizing the Pretraining Camera Positions}
\label{sec:appendix-scene-camera-coverage}
As discussed in ~\cref{sec:multi-scene-optimization}, we pretrain a view-mapper on a set of views that are consistent across scenes. We visualize this distribution in ~\cref{fig:camera-positions} (the figure is annotated with the sparse-view splits, but for pretraining we use all the visible cameras). This visualization was generated by running the introductory example for SDFStudio \cite{Yu2022SDFStudio} at their \href{https://github.com/autonomousvision/sdfstudio}{github repo}, which does visualization with the interactive nerfstudio viewer \cite{nerfstudio}. The two images are different views of the same 3D scene, but to get a better understanding of the 3D distribution, we recommend running the example SDFStudio code and interacting with the viewer directly. The cameras are equidistant from an origin that has the object (they live on a sphere), and they approximately point at the origin. They cover an azimuthal range around the origin of about 160$^\circ$, with a polar range of about 75$^\circ$. Note that although the cameras may be focused on one point, the object is not necessarily placed at that point, so the object will be not centered in all images (for example, refer to the ground truth images in ~\cref{fig:appendix-results-3nvs}).

\section{Visualizing the Camera Positions for DTU Sparse Dataset}
\label{sec:appendix-cameras-sparse-DTU}
In ~\cref{sec:results-multi-scene} and ~\cref{fig:multi-scene-concept}, we claimed that training on the 6-view split requires generalizing to some views that are `interpolations' and others that are `extrapolations'. To support this, we annotate the 6-view split that was used in ~\cref{fig:camera-positions} (we explain how we generated this visualization in \cref{sec:appendix-scene-camera-coverage}, and the explanation of training splits is in ~\cref{sec:appendix-eval}). All other visualized camera views are test views. Compared to the full set of images, the 6-view split covers most of the range in the azimuthal angle, $\varphi$, around the object, but only covers about half the range in the polar angle, $\theta$. It is intuitive that the test views outside the polar angle range are a more challenging generalization problem, and this is what we refer to as `extrapolation'. More concretely, the extrapolated views are those outside the convex hull of the train views in the spherical angle space $(\varphi,\theta)$.

\section{Pretraining Results}
\label{sec:appendix-pretraining-results}
\begin{figure*}
  \centering
    \includegraphics[width=\linewidth, page=7, trim={0 360 0 0}, clip]{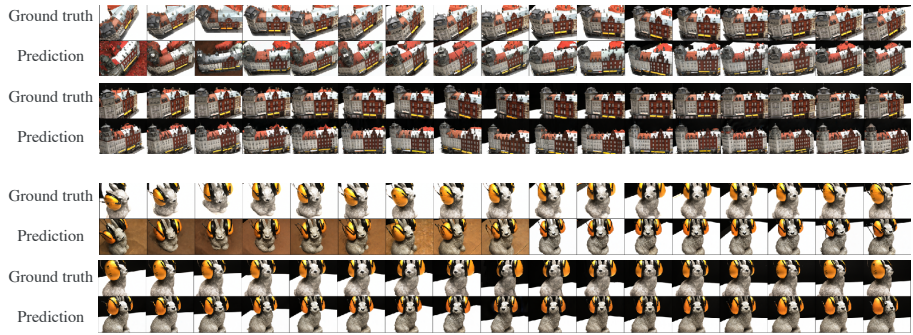}
  \captionof{figure}{Reconstructions of every training view in scenes 9 and 33 during pretraining for 100k steps. The scenes have the same view-mapper, but different scene-mappers. The correct reconstructions validates the view-mapper, which is shared across scenes, has learned the relative mapping between the camera viewpoints.}
  \label{fig:pretrain-recons}
\end{figure*}

In ~\cref{sec:single-scene-optimization}, we describe the model pretraining of dense views from multiple scenes from the DTU training dataset \cite{jensen2014large}. We use caption of the form ``$S_{R_i}$. A photo of an $S_{s_j}$''. The token for viewpoint, $S_{R_i}$, is controlled by the view-mapper and is shared across all scenes. The token for the $j$th scene, $S_{s_j}$, is controlled by an scene mapper and is the same for all views within a scene, but different across scenes. We verify that the view-mapper has learned the training views by checking reconstructions from sample scenes. In ~\cref{fig:pretrain-recons}, we reconstruct every view for the training scenes 9 and 33 after 100k steps of training

\begin{figure*}
  \centering
    \includegraphics[width=\linewidth, page=8, trim={0 0 890 0}, clip]{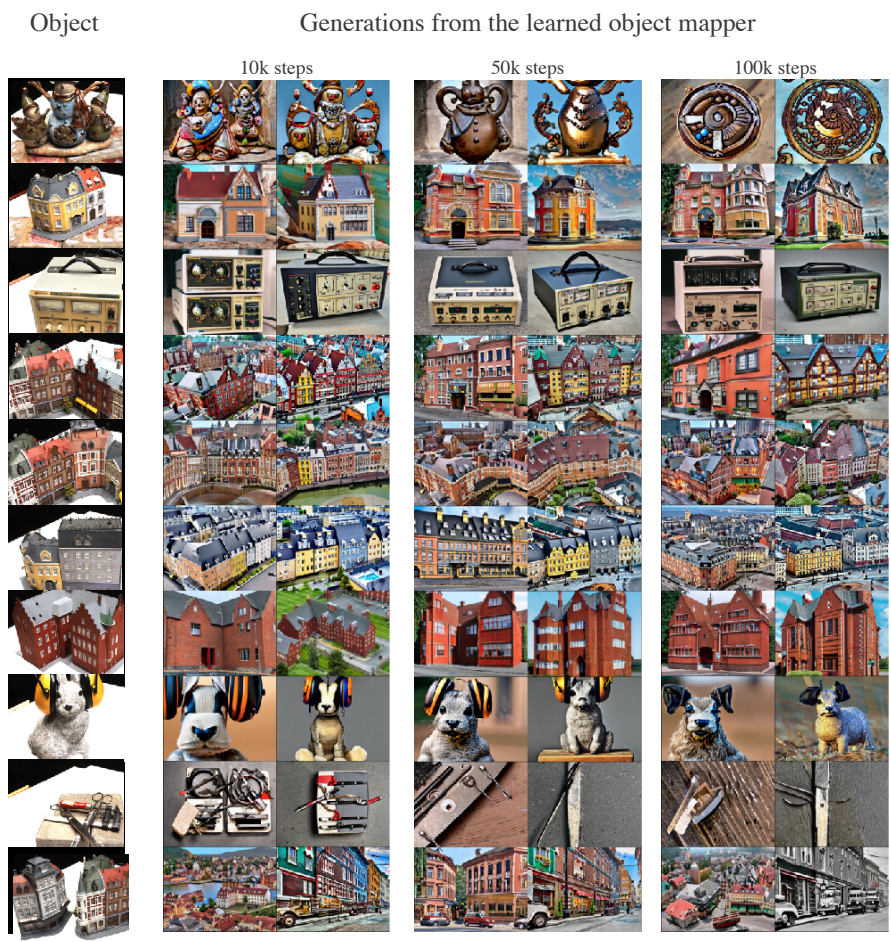}
  \captionof{figure}{Disentanglement results: for 10 random scenes in the pretraining dataset, we generate images for the scene-mapper without the view-mapper using prompts ``A photo of a $S_o$'' for special token $S_o$. The left column is view 25 from the original data. The other columns take two samples at different points in training.}
  \label{fig:disentanglement}
\end{figure*}

The captions above are constructed so that the scene token captures all the view-invariant scene semantics; that is, we want the viewpoint and semantics to be disentangled. One way to check this is to try generating images using only the scene token, e.g. `` A photo of an $S_{s_j}$''. We do this in ~\cref{fig:disentanglement} for a sample of the scene tokens at different points in training. This suggests the scene tokens approximately embed the scene semantics, though with varying degrees of success. Firstly, the generations are different and far more diverse than the original. Secondly, they have a different `style' to the DTU images; especially for the toy houses, they become more realistic, and arguably closer to the diffusion model training distribution. Third, for some scenes, the disentanglement become worse with more training time; in the first row, the generations have some similar attributes after 10k steps, but after 100k steps they are only a circular pattern.

In prior literature on textual inversion and personalization, evaluations are done on `reconstruction' ability or `fidelity' to the original images \cite{ruiz2023dreambooth, textual_inversion_gal2022, neti_alaluf2023neural}. This refers to the correctness of scene semantics and details in the generated images, and it tends to be traded off against `editability', which is the ability to compose the new concepts with existing concepts in the vocabulary \cite{neti_alaluf2023neural} by mixing the token with existing tokens in the prompt. How do these ideas relate to our setting? The reconstructions in the NVS experiments are good in most cases, and they are much more faithful to the images in the `disentanglement' test. We propose two explanations. First, the view-mapper could be embedding information not captured by the scene token; we would expect the view-mapper to capture attributes that are common across the scenes, and this might include common backgrounds (the table in DTU images is always white or brown, and there is a black background); or it could capture common `style' in the object semantics (e.g. there are many statues, many food items, and many toy houses). Second, it could be that generating images with only the scene token in the caption does not properly test disentanglement, but we do not have any further insights about this.

\section{Validation of Spherical Coordinate Parameterization}
\label{sec:appendix-spherical-coords}
\begin{figure*}
  \centering
    \includegraphics[width=\linewidth, page=5, trim={0 340 235 0}, clip]{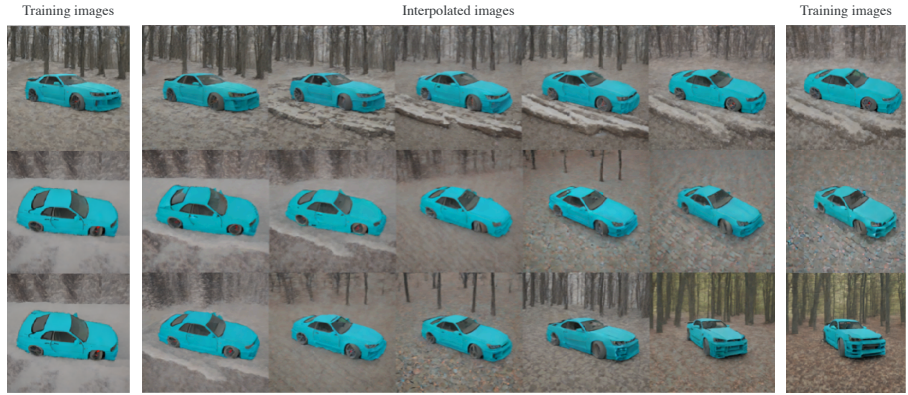}
  \captionof{figure}{Examples of novel view synthesis using ViewNeTI where the input camera parameters are in spherical coordinate system. We do single-scene NVS using an input dataset of nine multiview images of a ShapeNet car \cite{chang2015shapenet} with random forward-facing poses. The left and right columns are views in the multiview train set. The middle columns are NVS predictions from interpolating the polar and azimuth angles between the two edge images. The interpolation of the car poses are smooth and the scene semantic consistency is good..}
  \label{fig:spherical-coords}
\end{figure*}

For camera representation, our results use the camera-to-world projection matrix given in the DTU-MVS dataset \cite{jensen2014large}. But our method is agnostic to the camera parameterization. To show this, we show NVS results where the camera is parameterized by spherical coordinates in ~\cref{fig:spherical-coords}. We assume a central object is fixed at the origin, and that the camera is at a fixed radius with variable polar and azimuth angles, $(\theta, \varphi)$; we assume the camera is pointed at the origin. We do single-scene optimization of ViewNeTI on a rendering of a ShapeNet car \cite{chang2015shapenet}. To encourage the image to be close to the diffusion model training distribution, we generated an augmented training set by outfilling the background around the car, similar to in ~\cref{sec:appendix-infill-experiment}.  The left and right columns are reconstructions of camera poses from the multiview train set. The middle columns are NVS predictions from interpolating the polar and azimuth angles.

\section{Qualitative Comparison to Zero-1-to-3}
\label{sec:appendix-z123-baseline}
We discuss further the qualitative baseline comparison with Zero-1-to-3, \cite{liu2023zero1to3} from ~\cref{fig:baseline-comparison-single-view-nvs}. Zero-1-to-3 treats novel view synthesis as image-to-image translation with a diffusion model, conditioned on the change in camera parameter. It is trained on renderings from Objaverse, a large dataset of 3D assets available online \cite{deitke2023objaverse}. By training on a large dataset, Zero-1-to-3 is intended to  `zero-shot' generalize to new scenes without any extra finetuning, similar to zero-shot classification in CLIP \cite{clip_radford2021learning}. This has the advantage that NVS predictions are generated quickly - the time taken to generate a sample with Stable Diffusion model. On the other hand, this poses a very difficult challenge to generalize beyond the Objaverse data distribution. Unlike in CLIP, the training distribution of Zero-1-to-3 does not yet cover the full distribution of test scenes of interest for 3D perception applications \cite{jensen2014large, mildenhall2019local, dai2017scannet}, and this is because enormous multiview datasets are harder to collect that 2D image datasets. 


Finally, note that the failure modes of Zero-1-to-3 on DTU are distinct from the NeRF-based models in ~\cref{fig:baseline-comparison-single-view-nvs}, which all have imaging artifacts. Similar to ViewNeTI, the predictions do not have such artifacts, probably because both methods use diffusion models that are trained to generate images from a certain distribution of real images.

\section{ViewNeTI Implementation Details}
\label{sec:sup-details-architecure}
We use version \texttt{stabilityai/stable-diffusion-2-1} of  Stable Diffusion \cite{stable_diffusion_rombach2022high},  accessed from the diffusers library \cite{von2022diffusers}. The weights are all frozen. We did not test our method on Stable Diffusion1.

The inputs for camera parameters, timestep, and UNet layer are embedded to a 64-dim random Fourier feature vector. Specifically, for each of the 12 camera paramaters, one timestep, and one UNet layer, we sample 64 random frequencies from a standard normal, $\mathcal{N}(0,\sigma^2)$ where $\sigma$ is 0.5, 0.03, and 2 respectively. The encoding is computed as in \cite{tancik2020fourier2}, and as shown in our code. For the scene-mapper, the encoding is the same, but without the camera parameters. 

Following the base architecture of \cite{neti_alaluf2023neural}, the encoding is passed through an MLP with two blocks. Each block has a linear layer, LayerNorm \cite{ba2016layer}, and LeakyRelu \cite{xu2015empirical}. Finally, they are projected to 768 dimensions, which is twice the word-embedding for Stable Diffusion 2. This gives 140,000 parameters, which is the same for the view-mappers and scene-mappers. 

The word embedding input is scaled to have the same norm as a particular placeholder word, for example `statue' for the buddha statue scene (again, like in \cite{neti_alaluf2023neural}). We did one experiment on varying this word on one scene, and this showed that while norm scaling was important, the exact choice of word for the reference token was not, so we just used `object' for every scene in all experiments.

We use an effective batch size of 9 (batch size 3 with 3 gradient accumulation steps), a constant learning rate of 0.09, with the AdamW optimizer \cite{loshchilov2017decoupled} (again, like in \cite{neti_alaluf2023neural}). In training, DTU images were resized to (512, 384), which has the same aspect ratio as the DTU images. At inference, we found that image quality was better (fewer artifacts) if sampling at a higher resolution, (768, 576). Since Stable Diffusion 2 was trained with square images at (768,768), we experimented with padding DTU images to have the same aspect ratio, but we found these results to be worse.

\section{DTU Evaluation Details}
\label{sec:appendix-eval}
\begin{figure*}
  \centering    
    \includegraphics[width=0.7\linewidth, page=1, trim={0 0 1215 0},clip]{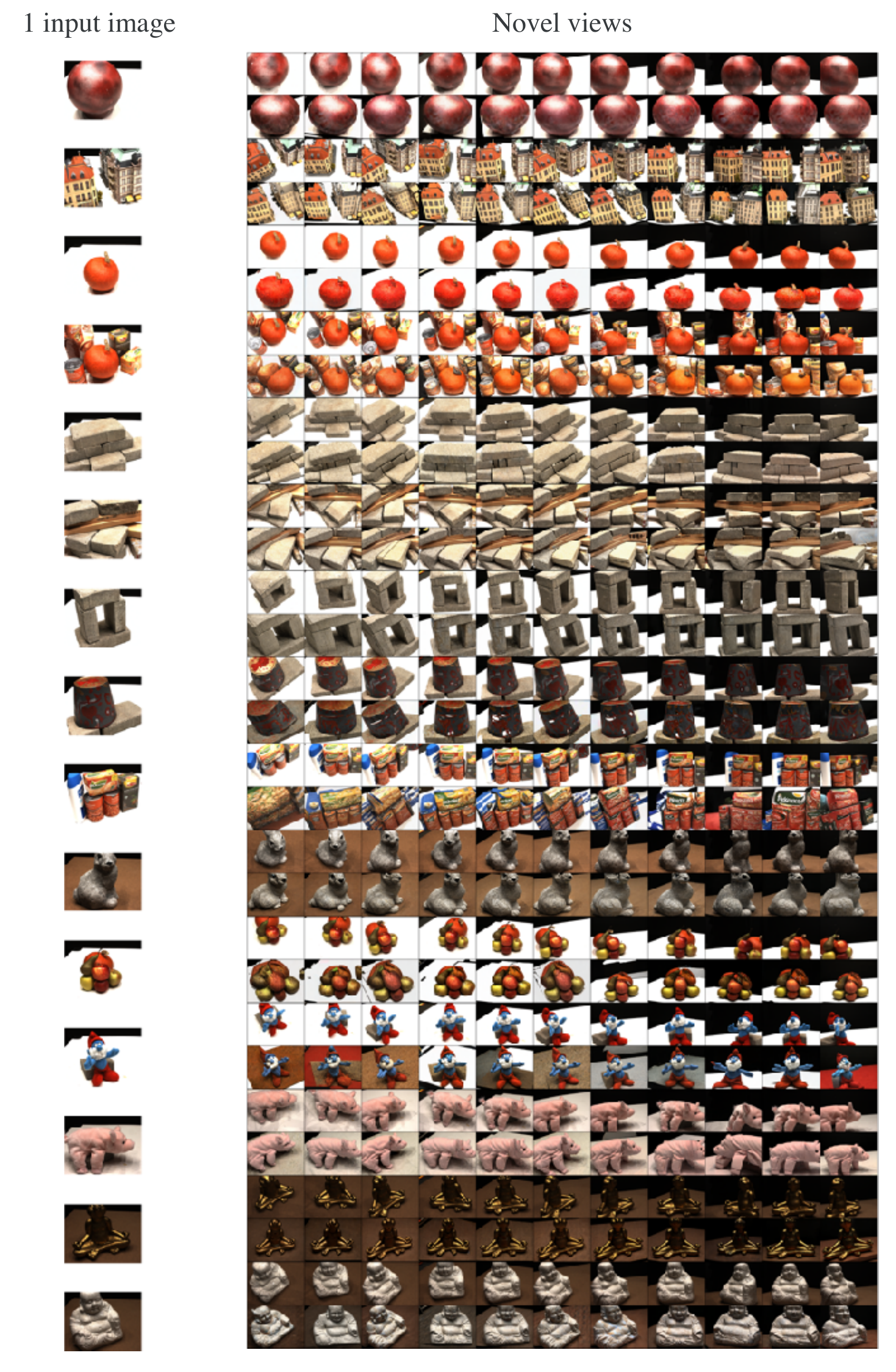}
  \captionof{figure}{ViewNeTI Novel view synthesis predictions for every DTU test set scene from one input view (the rows are alternating ground truth and prediction). All scenes have the same training hyperparameters and random seed  for training and generation. The hyperparameters are also the same as the three-view case in ~\cref{fig:appendix-results-3nvs}, except training steps are reduced from 3k to 1.5k, and the image augmentations are slightly changed as described in ~\cref{sec:appendix-data-augmentations}. In almost all cases, the rendered views are photorealistic, and the scene semantics are close to the ground truth, though semantics are worse for more complex scenes. The failure modes are incorrect scene details, and misaligned camera poses. By contrast, NeRF-based methods will have consistent scene semantics across views due to the explicit 3D representation, but have much worse image quality (see NeRF baseline comparison in ~\cref{fig:baseline-comparison-single-view-nvs} for these comparisons).  Different to the three-view case, another failure mode is for novel view predictions to be too close to the input view (overfitting). We  mitigate this by reducing the training steps, to 1.5k.  The views chosen for input are standard from previous work, and the novel viewpoints are chosen to cover the full sequence of views in DTU (see ~\cref{sec:appendix-eval}).  
  }
  \label{fig:appendix-results-1nvs}
\end{figure*}

\begin{figure*}
  \centering
    \includegraphics[width=0.7\linewidth, page=2, trim={0 0 1190 0},clip]{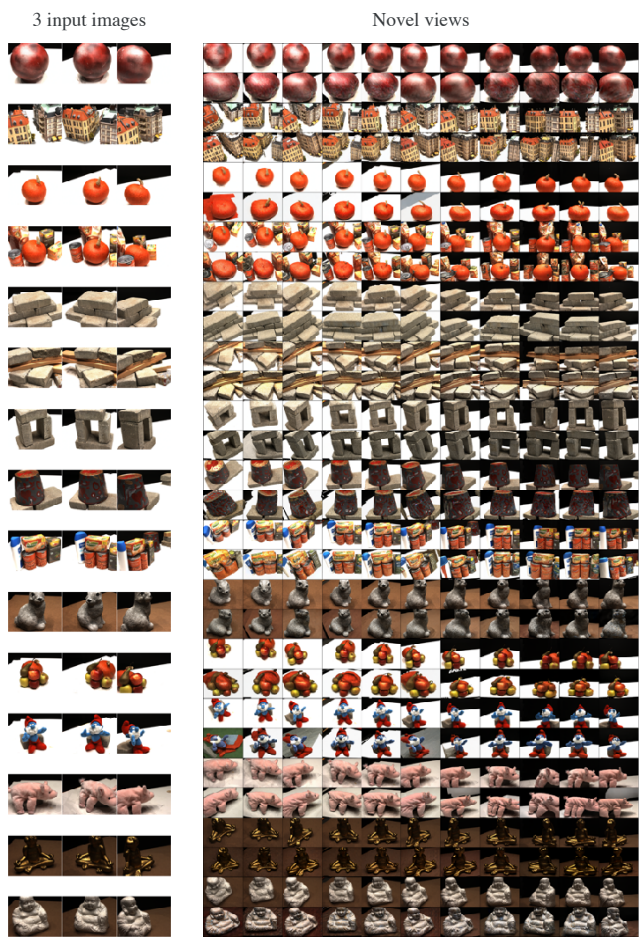}
  \captionof{figure}{ViewNeTI novel view synthesis predictions for every DTU test set scene from three input views (the rows are alternating ground truth and prediction). All scenes have the same training hyperparameters and random seed for training and generation. In almost all cases, the rendered views are photorealistic, and the scene semantics are close to the ground truth, though semantics are worse for more complex scenes. The failure modes are incorrect scene details, and misaligned camera poses. By contrast, NeRF-based methods have consistent scene semantics across views due to the explicit 3D representation, but much worse image quality (see NeRF baseline comparison in ~\cref{fig:baseline-comparison-single-view-nvs} for these comparisons). The views chosen for input are standard from previous work, and the novel viewpoints are chosen to cover the full sequence of views in DTU (see ~\cref{sec:appendix-eval}). 
  }
  \label{fig:appendix-results-3nvs}
\end{figure*}

The DTU \cite{jensen2014large} splits are the same as \cite{yu2021pixelnerf}. The test set scenes are (8, 21, 30, 31, 34, 38, 40, 41, 45, 55, 63, 82, 103, 110, 114), which are the ones visualized in ~\cref{fig:appendix-results-3nvs} and ~\cref{fig:appendix-results-1nvs}, and used for quantitative results in ~\cref{tab:results_dtu}. For pretraining, the train scenes are every non-test scene except (1, 2, 7, 25, 26, 27, 29, 39, 51, 54, 56, 57, 58, 73, 83, 111, 112, 113,115, 116, 117). They are excluded because they have too much overlap with the train set; e.g 115-117 have the same object as scene 114, but with a different pose. This ensures that there is a domain shift between train and test with respect to scene semantics.

The splits for views also follow prior work \cite{yu2021pixelnerf, niemeyer2022regnerf, yang2023freenerf, deng2023nerdi}. The index names for views are 0-indexed, while the DTU filenames are 1-indexed. The standard 9-view splits (which we do not experiment with here) use train views (25, 22, 28, 40, 44, 48, 0, 8, 13). The 6-view, 3-view, and 1-view splits choose the first 6, 3, and 1 views from that list respectively. The test views are all indexes from 0-49 except a set that were excluded due to image quality which are views (3, 4, 5, 6, 7, 16, 17, 18, 19, 20, 21, 36, 37, 38, 39). We use all the train and test views for pretraining, and unlike in PixelNerf \cite{yu2021pixelnerf}, we do not include the excluded views for pretraining.

In figures ~\cref{fig:appendix-results-1nvs} and ~\cref{fig:appendix-results-3nvs} we evaluate novel views in (1, 8, 12, 15, 24, 27, 29, 33, 40, 43, 48). We chose them as a representative subset of the test set views. In ~\cref{fig:baseline-comparison-single-view-nvs}, we evaluate the novel views (1, 45, 22), and they were chosen to match with evaluation in prior work \cite{deng2023nerdi}.

\section{Output Bypass Architecture for Scene Tokens}
\label{sec:appendix-output-bypass}
\begin{figure*}
  \centering
    \includegraphics[width=0.7\linewidth, page=9, trim={0 480 1200 0}, clip]{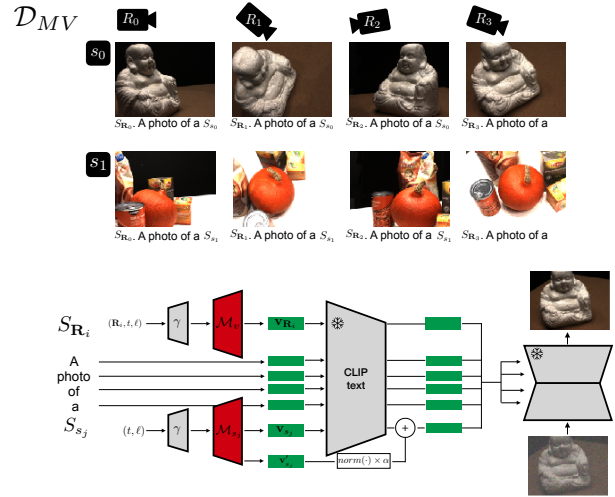}
  \captionof{figure}{
  As introduced in ~\cref{sec:appendix-output-bypass}, this is the modified system figure with an `output bypass' for the scene token, $S_{s_j}$. This is the extra branch from $\mathcal{M}_{s_j}$ that predicts $\mathbf{v}'_{s_j}$, which is normalized, multiplied by $\alpha$ and then added to the CLIP output for the scene token. 
  }
  \label{fig:output-bypass}
\end{figure*}

In \cref{sec:results-multi-scene} and \cref{fig:sysfig}, we describe multi-scene pretraining, and introduce the scene-mapper, $\mathcal{M}_{s_j}$ for predicting scene tokens of scene $s_j$. We now explain `output bypass', which is an implementation detail proposed by \cite{neti_alaluf2023neural} that modifies the standard textual inversion to improve the reconstruction of details in scenes. The changes to the system figure can be viewed in ~\cref{fig:output-bypass}. 

The scene-mapper is conditioned on diffusion timestep, $t$, and UNet layer, $\ell$, which are concatenated and passed through the Fourier feature encoding, $\mathbf{c}_\gamma=\gamma([t,\ell])$. Earlier, we stated that the scene-mapper follows the same equation as the view-mapper: it predicts a word embedding (the scene token):
\begin{equation}
    \mathbf{v}_{s_j} = \mathcal{M}_v(\mathbf{c}_\gamma)
\end{equation}

The dimension is the CLIP token dimension for that model, for example 768 in Stable Diffusion 2. When using an output bypass, we instead generate one extra `bypass' vector with the same dimension, $\mathbf{v}'_{s_j}$:
\begin{equation}
    (\mathbf{v}_{s_j}, \mathbf{v}'_{s_j}) = \mathcal{M}_v(\mathbf{c}_\gamma)
\end{equation}

To produce the extra vector, the MLP is exactly the same, but the output dimension is doubled, and then the output vector is chunked in two. Then, $\mathbf{v}'_{s_j}$ is scaled to have $L_2$-norm of $1$, and then multiplied by a scalar $\alpha$ which is set to 0.2 in our experiments. This vector is added to the original scene token \textit{after} it has been processed by the CLIP text encoder (see the ~\cref{fig:output-bypass}). 

The idea is that the output bypass can learn a small `perturbation' on the text encoder output. The choice of $\alpha$ ensures it cannot significantly change the text encoder output. In practice, it enables the scene token to learn finer-grained details without changing the coarse semantics \cite{neti_alaluf2023neural}.

\section{Pretraining ViewNeTI on Objaverse with inference on DTU}
\label{sec:appendix-viewneti-objaverse}
In the main results, we showed that after pretraining ViewNeTI on the 88 training scenes on DTU, we could then do single image NVS on test scenes from DTU. 
Although we argued that DTU test scenes were different object classes from DTU train scenes, reviewers were concerned that there may be a smaller distribution shift between DTU train and test, and therefore a comparison with Zero123 was unfair because it was pretrained on Objaverse. Originally we argued that Zero-1-to-3 pretraining on Objaverse is claimed to be general enough that inference on a new scene should work (it's `zero shot'). But to ameliorate this concern, we show that ViewNeTI can generalize from a small set of Objaverse as well (and that our performance is not due to test set similarity). Specifically, we pretrain ViewNeTI on 50 scenes from Objaverse and then perform single-image NVS on DTU test scenes. We compare that against two versions of Zero-1-to-3: one trained on the same 50 scenes, and the original model that is trained on 800,000 scenes. To test training on 50 scenes, we use the official repo \footnote{\tiny \url{https://github.com/cvlab-columbia/zero123}} and we verify that training reconstructions are correct.

\begin{figure}[h]
  \centering
  \includegraphics[width=0.85\columnwidth, page=12, trim={0 605 850 0}, clip]{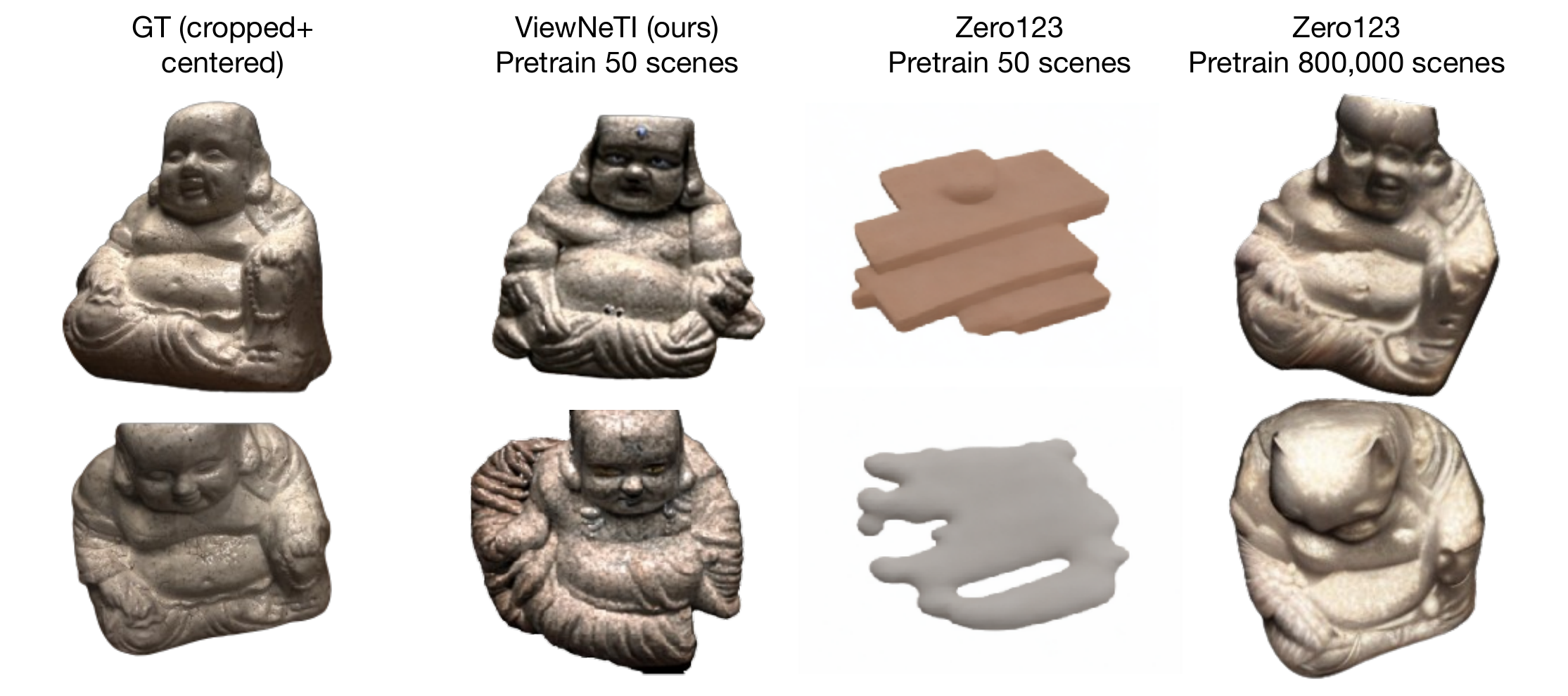}
   \caption{Single image NVS on DTU with pretraining on Objaverse, compared to Zero-1-to-3.}
   \label{fig:objaverse_training}
\end{figure}

The result is shown in \cref{fig:objaverse_training}. In this setting, ViewNeTI learns camera control, while Z123 on 50 scenes completely fails. The Z123 model trained on on 800,000 scenes learns some camera control, but the reconstruction quality is worse, based on qualitative observation. Also our LPIPS is 0.03 points better (0.37 vs 0.40) and  and PSNR is 0.2 points better (12.2 vs 12.0) compared to Z123 trained on 800,000 scenes.

\clearpage
\pagebreak

\begin{figure*}[t]
\centering
\small
\begin{lstlisting}[language=python]
    transform = T.Compose([
        T.RandomApply([T.RandomRotation(degrees=10, fill=1)], p=0.75),
        T.RandomResizedCrop(image_size, scale=(0.70, 1.3)),
        T.RandomApply([T.ColorJitter(0.04, 0.04, 0.04, 0.04)], p=0.75),
        T.RandomApply([T.GaussianBlur(5, (0.1, 2))], p=0.20),
    ])
\end{lstlisting}
\caption{PyTorch code for the image augmentations in single-view novel view synthesis. The 3-view NVS augmentations are slighty changed, as described in ~\cref{sec:appendix-data-augmentations}.}
\label{fig:augs}
\end{figure*}

\end{document}